\pdfoutput=1
\documentclass[sigconf,10pt]{acmart} %

\usepackage[utf8]{inputenc} 
\usepackage[T1]{fontenc}    
\usepackage{booktabs}       
\usepackage{nicefrac}       
\usepackage{microtype}      
\usepackage{amsmath}
\usepackage{amsfonts}       
\usepackage{graphicx,epstopdf}
\usepackage{tabularx}
\usepackage{subfigure}
\usepackage{enumitem}	
\usepackage{listings}
\usepackage{changepage}
\usepackage{pifont}
\usepackage{multirow}
\usepackage{wrapfig}
\usepackage{color}
\usepackage{tikz}
\usepackage[normalem]{ulem}
\usepackage{pifont}
\usepackage{soul}
\usepackage{xspace}
\usepackage{graphicx}
\usepackage{multirow}
\usepackage{multicol}
\usepackage{epsfig}
\usepackage{tikz}
\usepackage[toc,titletoc,page]{appendix}
\usepackage{graphicx}
\usepackage{varwidth}
\usepackage{capt-of}
\usepackage{comment}
\usepackage{bbm}
\usepackage{makecell}

\newcommand*\circleb[1]{\tikz[baseline=(char.base)]{
    \node[shape=circle,fill=black,draw,text=white,inner sep=1pt] (char) {#1};}}

\renewcommand{\sectionautorefname}{§\kern-2.5pt}
\renewcommand{\subsectionautorefname}{§\kern-2.5pt}
\renewcommand{\subsubsectionautorefname}{§\kern-2.5pt}

\makeatletter \DeclareRobustCommand\onedot{\futurelet\@let@token\@onedot}
\def\@onedot{\ifx\@let@token.\else.\null\fi\xspace}

\def\vs{\emph{vs}\onedot} 
\def\eg{\emph{e.g}\onedot} 
\def\ie{\emph{i.e}\onedot}

\def\etal{\emph{et al}\onedot}
\makeatother

\newenvironment{myitemize}%
  {\begin{itemize}
	[leftmargin=0cm,
		itemindent=.3cm,
		labelwidth=\itemindent,
		labelsep=0pt,
		parsep=3pt,
		topsep=2pt,
		itemsep=1pt,
		align=left]
  }%
  {\end{itemize}}

  {\begin{enumerate}
	[leftmargin=0cm,itemindent=.5cm,labelwidth=\itemindent,
		labelsep=0pt,
		parsep=1pt,
		topsep=1pt,
		itemsep=3pt,
		align=left]
  }%
  {\end{enumerate}}

\newenvironment{densedesc}{
\begin{description}[topsep=2.5pt, partopsep=0pt, leftmargin=1.5em]
  \setlength{\itemsep}{2.5pt} \setlength{\parskip}{0pt}
  \setlength{\parsep}{0pt}
}{\end{description}}

\newcommand{\runtime}{\textsf{Miro}\xspace}
\newcommand{\tx}{\textsf{TX2}\xspace}
\newcommand{\xavier}{\textsf{Xavier}\xspace}

\newcommand{\carm}{\textsf{CarM}\xspace}
\newcommand{\bsfull}{\textsf{BestStatic}\xspace}
\newcommand{\bs}{\textsf{BS}\xspace}
\newcommand{\bhfull}{\textsf{BestHistory}\xspace}
\newcommand{\bh}{\textsf{BH}\xspace}
\newcommand{\heufull}{\textsf{Heuristic}\xspace}
\newcommand{\heu}{\textsf{Heu}\xspace}

\usepackage{arydshln}
\makeatletter
\def\adl@drawiv#1#2#3{%
        \hskip.5\tabcolsep
        \xleaders#3{#2.5\@tempdimb #1{1}#2.5\@tempdimb}%
                #2\z@ plus1fil minus1fil\relax
        \hskip.5\tabcolsep}
\newcommand{\cdashlinelr}[1]{%
  \noalign{\vskip\aboverulesep
           \global\let\@dashdrawstore\adl@draw
           \global\let\adl@draw\adl@drawiv}
  \cdashline{#1}
  \noalign{\global\let\adl@draw\@dashdrawstore
           \vskip\belowrulesep}}
\makeatother

\newcommand\Authfont{}
\newcommand\Affifont{}

\usepackage{mathtools}
\usepackage{threeparttable}
\usepackage{balance}
\usepackage[algoruled,boxed,lined]{algorithm2e}
\usepackage[font=small,skip=2pt]{caption}

\usepackage{caption}
\usepackage{subfigure}
\newcommand{\parlabel}[1]{\vspace{0.2em}\noindent\textbf{#1}}

\setcopyright{acmcopyright}

\begin{document}

\author{\Authfont{Xinyue Ma}}
\affiliation{%
  \institution{\Affifont{UNIST}}
  \city{}
  \country{}
}

\author{\Authfont{Suyeon Jeong}}
\affiliation{%
  \institution{\Affifont{UNIST}}
  \city{}
  \country{}
}

\author{\Authfont{Minjia Zhang}}
\affiliation{%
  \institution{\Affifont{Microsoft}}
  \city{}
  \country{}
}

\author{\Authfont{Di Wang}}
\affiliation{%
  \institution{\Affifont{Unaffiliated}}
  \city{}
  \country{}
}

\author{\Authfont{Jonghyun Choi}}
\affiliation{%
  \institution{\Affifont{Yonsei University}}
  \city{}
  \country{}
}

\author{\Authfont{Myeongjae Jeon}}
\affiliation{%
  \institution{\Affifont{UNIST}}
  \city{}
  \country{}
}


\acmYear{2023}\copyrightyear{2023}
\setcopyright{acmlicensed}
\acmConference[ACM MobiCom '23]{The 29th Annual International Conference on Mobile Computing and Networking}{October 2--6, 2023}{Madrid, Spain}
\acmBooktitle{The 29th Annual International Conference on Mobile Computing and Networking (ACM MobiCom '23), October 2--6, 2023, Madrid, Spain}
\acmPrice{15.00}
\acmDOI{10.1145/3570361.3613297}
\acmISBN{978-1-4503-9990-6/23/10}

\ccsdesc[500]{Computing methodologies~Machine learning}
\ccsdesc[500]{Human-centered computing~Ubiquitous and mobile computing}

\keywords{Continual learning, On-device Training, Energy Efficiency, Episodic Memory Management }

\title{Cost-effective On-device Continual Learning over Memory Hierarchy with \runtime}

\begin{abstract}

Continual learning (CL) trains NN models incrementally from a continuous
stream of tasks. To remember previously learned knowledge, prior studies
store old samples over a memory hierarchy and replay them when new tasks
arrive. Edge devices that adopt CL to preserve data privacy are typically
energy-sensitive and thus require high model accuracy while not
compromising energy efficiency, i.e., cost-effectiveness. Our work is the
first to explore the design space of hierarchical memory replay-based CL to
gain insights into achieving cost-effectiveness on edge devices. We present
\runtime, a novel system runtime that carefully integrates our insights
into the CL framework by enabling it to dynamically configure the CL system
based on resource states for the best cost-effectiveness. To reach this
goal, \runtime also performs online profiling on parameters with clear
accuracy-energy trade-offs and adapts to optimal values with low overhead.
Extensive evaluations show that \runtime significantly outperforms baseline
systems we build for comparison, consistently achieving higher
cost-effectiveness.

%

\end{abstract}

\maketitle
\renewcommand{\shortauthors}{Xinyue Ma, Suyeon Jeong, Minjia Zhang, Di Wang, Jonghyun Choi, Myeongjae Jeon}
\section{Introduction}

\begin{figure}[!t]
\center
\includegraphics[width=\linewidth]{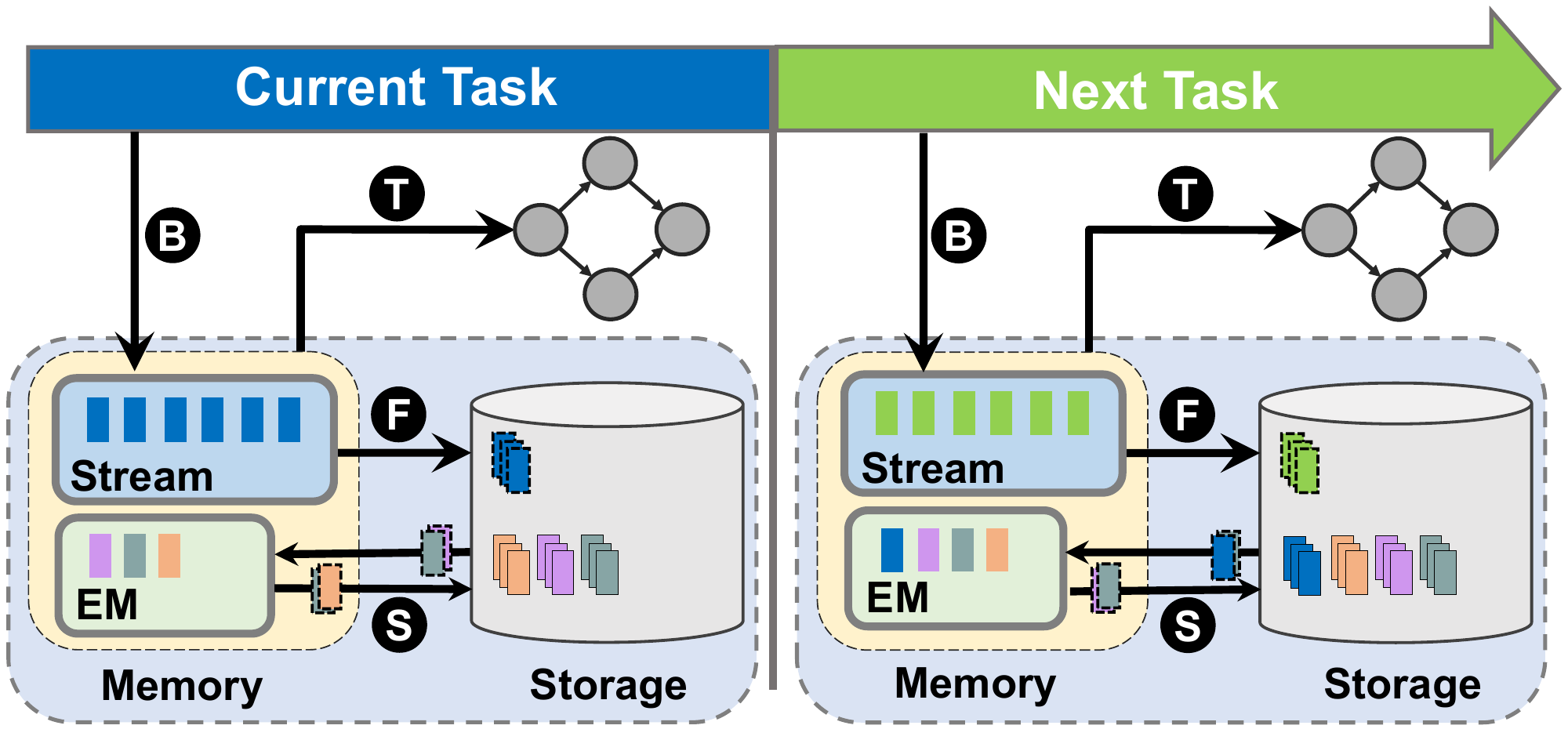}
\vspace{-0.1in}
\caption{Architecture and execution stages of HEM.}
\vspace{-0.1in}
\label{fig:arch_2}
\end{figure}

As the demand for realistic on-device machine learning (ML) has grown,
learning paradigms well-suited for mobile and edge devices have attracted
significant attention. One such paradigm is continual learning (CL), which
performs model training \emph{incrementally} as new data
becomes available. The neural network (NN) models in CL running on devices
learn new knowledge from data without dependencies on scalable compute
resources provided by external servers, preserving strong data privacy. As
\emph{on-device CL} research is getting mature, particularly concerning its
system and data efficiency~\cite{sparcl, hayes2022online, kwon2021exploring},
we anticipate it to gain prominence in many privacy-sensitive tasks on the
edge, such as user personalization~\cite{pellegrini2021continual,
diwan2022continual, chauhan2020contauth},
home automation~\cite{e-domainil, pellegrini2021continual}, smart
healthcare~\cite{vokinger2021continual, leite2022resource}, wearable
devices~\cite{schiemer4357622online}, and video analytics~\cite{ekya, recl,
shaheen2022continual}.

However, any CL algorithms, including on-device CL, must overcome
\emph{catastrophic forgetting} to realize their full
potential~\cite{mccloskeyC89}. CL undergoing severe forgetting will have the
previously learned knowledge quickly fading away
while learning new data. ML researchers have extensively explored this issue
and episodic memory (EM) has been recognized as one of the most effective
approaches~\cite{darker, AGEM, tiny, LopezPaz2017GradientEM, gdumb}. With EM,
CL stores old samples in the memory and \emph{replays} them along with new
samples during training to reduce forgetting.
Edge devices typically have limited memory capacity (4--12~GB) for all
on-device programs,
making it difficult to dedicate ample memory space solely for EM although
large EM promises high performance. Consequently, researchers have begun
exploring
a pragmatic system design for \underline{EM} over a deep memory
\underline{H}ierarchy (\underline{HEM})~\cite{carm}, including small memory
with fast access (50–150~ns) and large storage with slow access
(25–250~$\mu$s).

\parlabel{EM over Memory Hierarchy (HEM).}
The HEM builds upon the existing EM-based CL methods, as shown
in~\autoref{fig:arch_2}, while considering both system efficiency and
prediction accuracy. HEM organizes EM in RAM to retrieve a small set of old
samples at high speed during training. At the same time, HEM stores a large
number of old samples in storage and uses them to improve model accuracy by
performing \emph{data swapping}.
The data swapping is an online process that replaces the samples in the EM
drawn by model training (\ie, old samples replayed to mitigate the
forgetting) with other old samples in storage.
With HEM, CL can constantly retrieve information from a large corpus of past
data to memorize it
without increasing model capacity.

\begin{table}[]
 \begin{tabular}{@{}l@{}}
 \specialrule{1.8pt}{0.7pt}{0.7pt}

\small
 \makecell[l]{1. GPU dominates dynamic power. Thus, perform
 data swapping \\ as needed without concern for its impact on power usage.} \\
 \midrule

\small
 \makecell[l]{2. Using larger memory for new and old tasks often harms cost-\\
 effectiveness. No silver-bullet solution to memory allocation exists. \\
 Thus, make informed dynamic decisions for the current CL setup.} \\
 \midrule

\small
 \makecell[l]{3. The epoch count required to achieve near-highest accuracy \\
 is observed to remain consistent over consecutive training tasks. \\
 Attempts to reduce this count may have negative consequences. \\
 So, use the known good value or search it for the first few tasks.} \\
 \specialrule{1.8pt}{0.7pt}{0.7pt}

 \end{tabular}
\centering
\caption{Key insights from our study of on-device CL.}
\label{tab:insights}
\vspace{-0.2in}
\end{table}

HEM is suited to hardware requirements of edge computing systems for several
reasons. (1) Removable storage devices are cheap. For example, one of our
reference platforms, NVIDIA Jetson TX2, is equipped with 8~GB RAM and 32~GB
eMMC flash, and its storage can be easily extended with a 256~GB microSD card
at only around 20~USD. (2) Sustainable I/O traffic is low. The eMMC flash
drive in Jetson TX2 serves I/O requests at around 400~MB/s (and
$\sim$100~MB/s for microSD); in comparison, NVMe SSDs in modern AI servers
support I/O rates up to 3.5 GB/s. However, by swapping only a small number of
EM samples, it is possible for HEM to
match those low I/O bandwidth storage. (3) Processing capabilities of mobile
GPUs are not powerful enough. HEM's key advantage lies in enhancing the data
diversity of traditional EM-based CL methods without inducing additional
computational costs, which can hamper edge AI GPUs and manifest as longer
training times.

\parlabel{Efficient and Practical HEM.}
Recent studies like CarM~\cite{carm} have demonstrated that HEM can
remarkably improve model accuracy in image classification. However,
evaluating HEM has been limited to a proof-of-concept and overlooked the most
important metric of edge devices, \emph{energy efficiency}. Moreover, CarM
conducted experiments focusing on a few
swapping ratios on an AI server with hardware specifications vastly different
from ordinary edge devices. These swapping ratios were neither systematically
chosen nor based on the consideration of energy spending associated with EM
size and data movements between storage and memory.

In this work, we take a step toward building more practical on-device
CL by exploring the design space of HEM using various edge platforms and
workloads and providing valuable insights with the real-world evaluation
metric of energy consumption. Also, we showcase a new system runtime that
optimizes HEM for high \emph{cost-effectiveness}, which achieves high
prediction accuracy with lower energy consumption, under diverse resource
constraints of target devices.

\parlabel{Contributions.}
We make the following contributions:

$\bullet$ \emph{Systematic study of on-device CL.} To the best of our
knowledge, this work presents the most comprehensive exploration of on-device
CL over memory hierarchy to date. We compare the performance, accuracy, and
energy efficiency of various points in the design space of HEM, as listed
in~\autoref{tab:parameters}.
Our study spans two popular NN models and two CL methods designed for
Computer Vision (CV).
Our study provides key insights as summarized in~\autoref{tab:insights}.

$\bullet$ \emph{System runtime for HEM.}
We develop \runtime, a novel system runtime that optimizes HEM for
cost-effectiveness based on the insights we have gained from our exploration
study. A key challenge is configuring the various design parameters
in~\autoref{tab:parameters} without compromising accuracy and energy
efficiency. To address this challenge, we categorize these parameters into
three types: \emph{Static}, \emph{Capacity}, and \emph{Trade-off}, and employ
different optimization techniques for each type. For example, the data swap
ratio is a capacity-bound parameter with minimal impact on system-wide energy
usage. Hence, \runtime only controls I/O congestion and attempts to make the
most of the available I/O bandwidth. In contrast, \runtime dynamically
determines the optimal memory space for new and old tasks while considering
their accuracy-energy trade-offs, which can vary significantly over time. To
make informed choices, \runtime estimates accuracies and energy expenses of
potential size pairs via online profiling, identifies promising
configurations, and applies a cost-effectiveness metric called \emph{utility}
to select the
best candidate. We propose several cost-reduction techniques to make
profiling more feasible.

\parlabel{Experimental Results.}
We build \runtime atop the state-of-the-art HEM-based CL system CarM and
evaluate its efficacy against baseline systems we implement for comparison on
five datasets across three task domains, including image classification,
audio classification, and human activity recognition, as shown
in~\autoref{tab:datasets}. In small to medium-scale experiments, \runtime
consumes 7--66\% less energy while achieving 1.35--15.37\% higher accuracy
across various memory budgets compared to the baseline systems. In
large-scale experiments with the ImageNet1k dataset~\cite{imagenet_1k},
\runtime achieves 46.05\% energy savings and 23.37\% higher accuracy compared
to CarM. Due to our cost-saving efforts, our profiler is 346$\times$ faster
than an exhaustive one that uses all epochs and samples and adds only around
7--10\% overhead, which can easily be offset by the benefits that \runtime
delivers.

\section{Continual Learning on HEM}
\label{sec:background}

In continual learning (CL), the NN model is trained with a stream of non-IID
data. This input data is typically organized as multiple tasks, where a task
may include data that diverges from past data, referred to as
\emph{data drift}. Examples include (1) unseen object appearances, scenes,
and lighting conditions for video analytics~\cite{recl, ekya}, (2) new
activities and gestures for human activity
recognition~\cite{kwon2021exploring}, and (3) new identity of the individuals
and pets for household robots~\cite{hayes2022online, e-domainil}.

These scenarios often necessitate timely outcomes from training. For
instance, consider anomaly detection in video stream analytics that aims to
identify abnormal objects or actions in images captured by surveillance
cameras, drones, or robots. In such cases, an initially deployed model cannot
identify all possible abnormal events in advance. Since anomalies demand
immediate attention, updating models promptly is essential: postponing
training to unpredictable intervals (e.g., device charging) would limit the
applicability of CL. Moreover, not all mobile and edge devices have
sufficient memory to accommodate training data due to heterogeneous hardware
resource constraints. The purpose of \runtime is to tackle these intricate
challenges and broaden the scope of scenarios to which CL can be effectively
applied.

\parlabel{Workflow.} In edge devices, storage can be used
alternative to memory
to store and replay a large amount of data from past tasks to help maintain
high accuracy even with data drift.
However, fundamentally, single-level non-hierarchical EM design cannot have the best
of both worlds, i.e., high speed and ample space~\cite{carm}. HEM addresses
the limitations of single-level EM by optimizing it over a deep memory
hierarchy. \autoref{fig:arch_2} shows the execution stages a new task goes
through in HEM.

\begin{myitemize}

\item \textbf{Buffering} \circleb{\textsf{B}}:
    The samples for a new task $N$ are accumulated
    in a \emph{stream buffer} (SB) and later retrieved for training. The size
    of the task $N$ depends on the learning method in use, which can be either
    large new class data~\cite{icarl, bic} or simply a small chunk of new
    samples~\cite{tiny, aser}.

\item \textbf{Training} \circleb{\textsf{T}}:
    HEM combines samples from the task $N$ in the
    SB with old samples in EM to compose training data. These
    combined samples are organized into mini-batches, which are then fed into
    the model one at a time, constituting training iterations. This
    process can repeat multiple rounds, known as epochs, to improve
    model accuracy.

\item \textbf{Swapping} \circleb{\textsf{S}}:
    During the training of the task $N$, HEM swaps
    data between in-memory samples and in-storage samples of the same class.
    This is performed asynchronously to prevent slow storage access from
    delaying the continual model training. In cases where the available I/O
    bandwidth is insufficient, only a small subset of EM samples are swapped.
    By default, HEM randomly selects these samples~\cite{carm}.

\item \textbf{Flushing} \circleb{\textsf{F}}:
    Once the training of the task $N$ is completed, 
    EM is updated with the samples in the SB.
    If EM is short of memory space,
    HEM applies a
    sampling strategy to reorganize EM with a subset of the task's samples.
    Old tasks should evict some samples to avoid the overflow of EM.
    The stream data is then flushed onto the storage and cleaned up for the
    next incoming task $N+1$.
    We keep both in-memory and in-storage data class-balanced within the capacity.

\end{myitemize}

\begin{figure}[!t]
\centering
    \includegraphics[width=\linewidth]{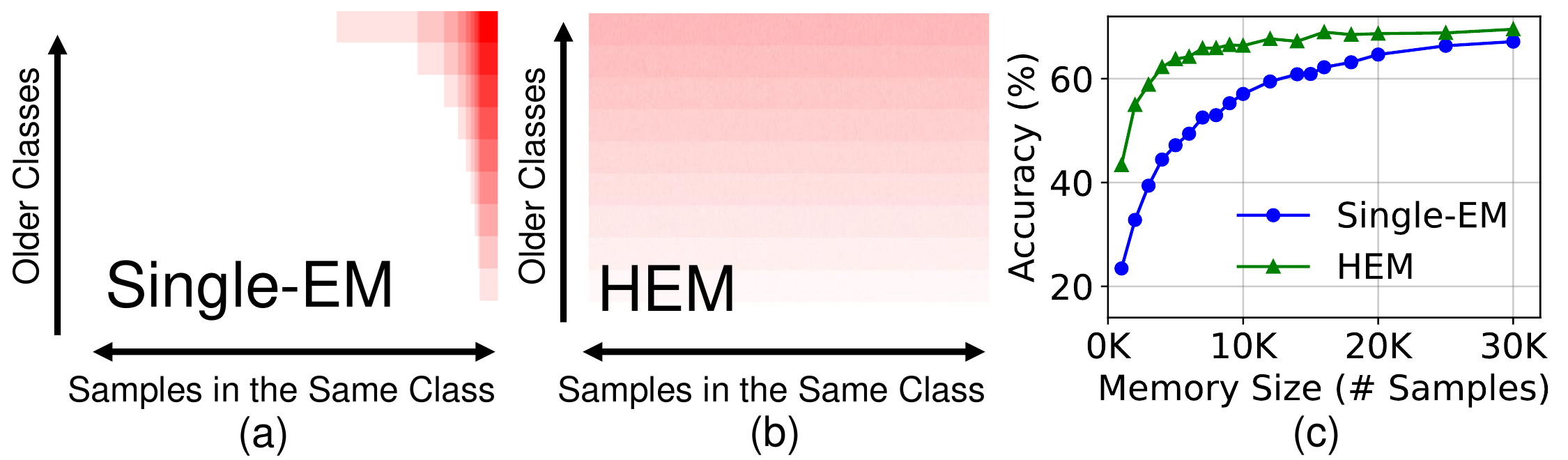}
    \caption{(a) and (b): Data diversity of old samples between single-level EM and HEM. (c): Accuracies over memory sizes.}
    \label{fig:diversity}
\vspace{-0.2in}
\end{figure}

\parlabel{Data Diversity.}
In HEM,
any samples in storage have an opportunity to be replayed in the future.
More diverse old samples can help reduce forgetting of their tasks. \autoref{fig:diversity}(a) and (b) illustrate the
range of old samples that take part in training between 
single-level EM design and HEM, respectively, over ten tasks that contain
different classes from the CIFAR100 dataset~\cite{cifar}.
\autoref{fig:diversity}(c) compares the final top-1 accuracy averaged across
all classes between the two methods for image classification using the
ResNet-32 model~\cite{resnet}.
\section{Design Space Evaluation}
\label{sec:characterization}

In this section, we analyze the design parameters of HEM to gain a thorough
understanding of their impact on both system and model efficiency.
Traditionally, these parameters have been configured \emph{statically} to
achieve high model accuracy for non-hierarchical EM designs~\cite{ercl, icarl,
bic, tiny, gdumb, darker, rainbow}. However, simply repurposing this static
configuration for HEM on edge devices without considering its impact on
energy efficiency would not yield practical benefits. Through our study, we
offer valuable insights to aid the development of a system runtime for
cost-effective HEM.
\begin{table*}[]
 \begin{tabular}{@{}lll|ll@{}}
 \multicolumn{1}{l}{\textsf{\textbf{Parameter}}}
 & \multicolumn{1}{l}{\textsf{\textbf{Description}}}
 & \multicolumn{1}{l|}{\textsf{\textbf{Resource}}}
 & \multicolumn{1}{l}{\textsf{\textbf{Decision}}}
 & \multicolumn{1}{l}{\textsf{\textbf{Constraint}}}\\
 \specialrule{1.8pt}{0.7pt}{0.7pt}

 	EM size
	& \# old samples populated in RAM
    & Memory
    & Dynamic
    & Trade-off \\

 	SB size
	& \# new task samples to keep in RAM for training
    & Memory
    & Dynamic
    & Trade-off \\

 	Swap ratio
    & Average \% for EM samples to be replaced by other samples in storage
    & I/O
    & Dynamic
    & Capacity \\

 	Epoch count
    & \# times the data combining EM and SB is trained upon a new task
    & GPU
    & Static
    & Static \\

 \specialrule{1.8pt}{0.7pt}{0.7pt}

 \end{tabular}
\centering
\caption{Descriptions for four design parameters of HEM. For each parameter, we show the HW resource type it exercises, how to make configuration decisions, and the constraint to which it is bounded.
}
\label{tab:parameters}
\vspace{-0.2in}
\end{table*}

\subsection{Methodology}
\label{sec:methodology}

\parlabel{System Platform.}
Our study utilizes NVIDIA Jetson TX2 (\tx)~\cite{jetson_tx2} and Jetson
Xavier NX (\xavier)~\cite{jetson_xavier_nx} as reference edge devices. \tx is
equipped with a 256-core NVIDIA Pascal GPU, 8GB of RAM, and a 32GB eMMC 5.1
drive. \xavier is equipped with 384-core NVIDIA Volta GPU, 16GB of RAM, and a
16GB eMMC 5.1 drive. The RAM is a single unified memory shared by the CPU and
GPU.
To conduct experiments, we adopt CarM~\cite{carm_code}, which runs on PyTorch
1.7.1 and has been enhanced with the ability to dynamically configure HEM
parameters.

\parlabel{Datasets.}
\autoref{tab:datasets} provides a summary of the full datasets used in
evaluating \runtime. However, for this particular characterization study, we
focus on image classification tasks and utilize two popular datasets:
CIFAR100~\cite{cifar} (100 classes, 197MB) and
Tiny-ImageNet~\cite{TinyImageNet} (200 classes, 1.1GB). By default, we split
each dataset into 10 tasks, with each task comprising of 10 classes for
CIFAR100 and 20 classes for Tiny-ImageNet, guided by many prior works in CL
domain~\cite{icarl, bic, gdumb, darker, rainbow}. These tasks are issued
sequentially and do not overlap, ensuring that each sample appears only once.

While these datasets can fit easily in the RAM of \tx and \xavier, it is
important to note that on-device CL is restricted by energy expenses incurred
by training in-memory data (i.e., SB plus EM), rather than by
memory usage to maintain the data. In fact, the impact of memory usage is a
double-edged sword. For instance, a larger EM may improve accuracy, but it
can also increase energy consumption due to the larger training data, often resulting in diminishing returns or no accuracy improvement compared to smaller EMs.
Although it is worthy of evaluating gigantic datasets like
ImageNet1k~\cite{imagenet_1k} (146GB)
to demonstrate longer-term effects (\autoref{sec:large-scale-eval}), we
urge caution against always \emph{using up} all available memory for sizing
SB and EM, irrespective of the dataset in use.

\parlabel{Training Methods.}
We employ ResNet-32 and ResNet-18 as the reference models for CIFAR100 and
Tiny-ImageNet, respectively. To run these models on HEM, we have chosen two
CL methods: ER~\cite{ercl} and BiC~\cite{bic}. ER is a pure episodic
memory-based method without any additional algorithmic optimization, whereas
BiC incorporates \emph{knowledge distillation}, a widely used CL technique
that transfers learned knowledge backward as more tasks are added.

\parlabel{Metrics.}
To measure power consumption, we exploit the built-in sensor values for the
GPU, RAM, CPU, and I/O connection between the RAM and storage on the Jetson
device~\cite{power_manual, power_code}.
Power is measured in watts (W) and energy is gauged in joules (J) by
multiplying power by time. As for the accuracy metric, we calculate the final
accuracy averaged across all classes after completing the last task training.
Final average accuracy has been a standard metric used in CL
to measure accuracy over consecutive task insertions~\cite{icarl, bic,
gdumb, darker, rainbow}.

\subsection{Design Parameters}
\label{sec:parameters}

\autoref{tab:parameters} outlines four representative parameters -- \emph{EM
size}, \emph{SB size}, \emph{swap ratio}, and \emph{epoch
count} -- that impact the performance, accuracy, and energy efficiency of
HEM.
These parameters capture the resource usage required for new task training.
In general, higher resource usage leads to higher accuracy. However, each
parameter has unique implications on cost-effectiveness, which refers to the
accuracy improvement achieved for the energy spent. For example, more epochs
consume energy proportionally but only offer minimal accuracy improvement
once the model converges, resulting in low cost-effectiveness.

It is worth noting that each parameter exercises specific HW resources within
the system. EM and SB operate on RAM, data swapping mainly involves I/O
operations, and epoch count pertains to GPU usage for model training. This
also means that certain parameters may compete for the same resources when
capacity constraints arise. For instance,
the GPU is exclusively dedicated to model training,
whereas I/O bandwidth is a system resource that varies in its availability as
all on-device programs carry out I/O operations. Our system runtime will be
designed to adapt to these interferences based on the observations made in
this study.

\begin{figure}
    \centering
    \includegraphics[width=\linewidth]{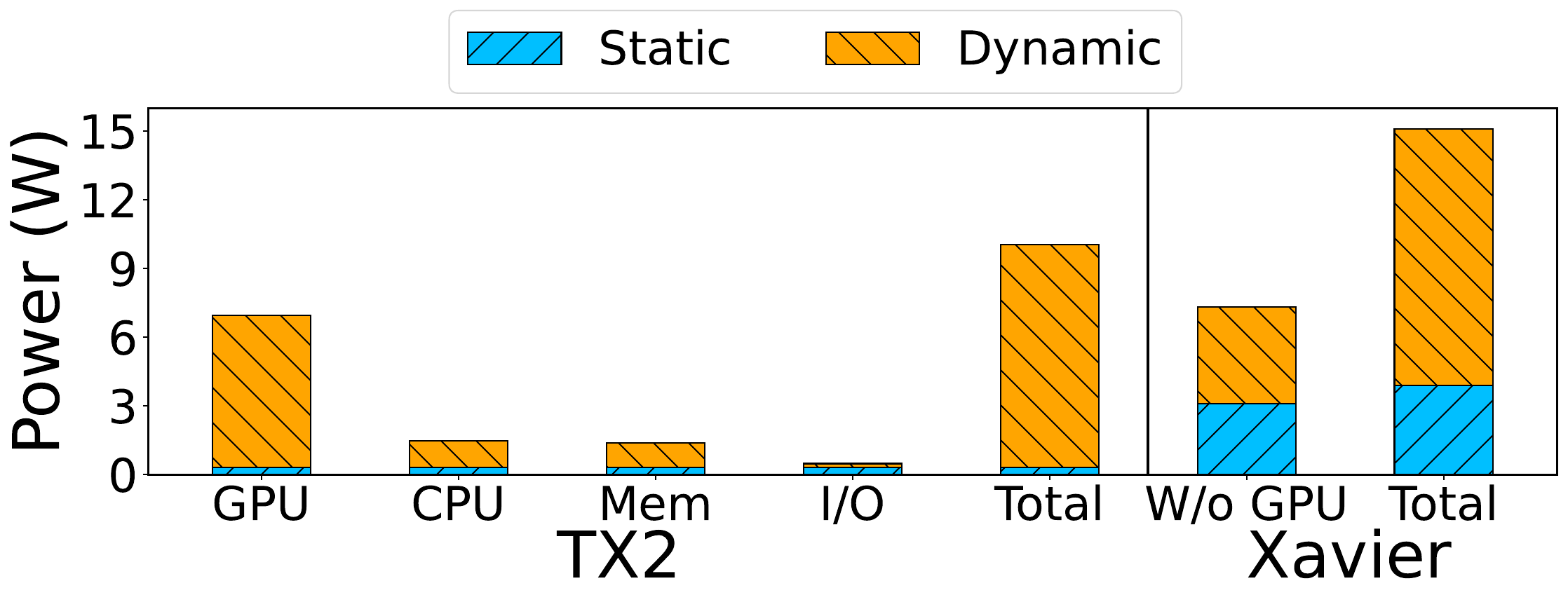}
    \vspace{-0.2in}
    \caption{Power consumption of HEM across major system components on NVIDIA Jetson TX2 and Jetson Xavier NX.}
    \vspace{-0.2in}
    \label{fig:power_consumptions}
\end{figure}

\begin{figure*}[!t]
\centering
    \includegraphics[width=\textwidth]{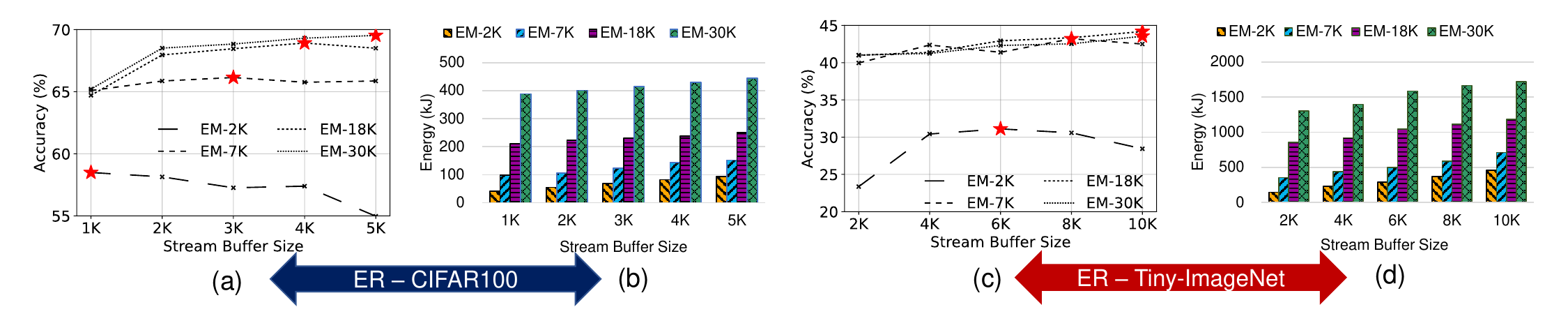}
    \vspace{-0.2in}
\caption{(a) and (c): Accuracy over varying SB sizes and EM sizes in terms of the number of samples for CIFAR100 and Tiny-ImageNet. (b) and (d): Their corresponding energy consumption.}
\vspace{-0.1in}
\label{fig:memory_params_tiny}
\end{figure*}

\subsection{Energy-Accuracy Characteristics}
\label{sec:exploration_results}

\subsubsection{Impact on System Power}
\label{subsec:power_impact}

Compared to other methods, HEM exercises more system resources: \eg,
traditional EM methods do not operate I/O for data swapping and non-EM
methods do not even manage memory for replaying. Therefore, it is crucial to
understand the power characteristics of HEM
that arise from changing parameter values.

For this analysis, we run the HEM using CIFAR100 and ER with full data
swapping, \ie, 100\% swap ratio, on both \tx and \xavier, using default power
modes with a power budget of around 10W for \tx and 15W for \xavier.
\autoref{fig:power_consumptions} illustrates power consumption by
distinguishing between static and dynamic power to precisely confirm the
impact of training. We make three observations. (1) Dynamic power dominates,
resulting in the total power consumption nearly reaching the power budget for
each device. (2) Among system components, the GPU is responsible for most
dynamic power consumption. While \tx reports the GPU power separately,
\xavier reports a single number for all on-chip components. To determine the
effect of GPU dynamic power, we make the GPU consume only static power by
inactivating it during the training process, which is shown in the
\textsf{w/o GPU}. We found that the GPU dynamic power still accounts for up
to 42.7\% in Xavier. (3) I/O activity driven by data swapping accounts for a
small portion of the power consumption ($\sim$0.1W). Even when we saturate
I/O bandwidth, I/O power usage does not increase its portion significantly.
Our findings are summarized in the first row of~\autoref{tab:insights}.

\subsubsection{Design Parameters}
\label{subsec:design parameters}

\parlabel{Memory Parameters: EM and SB.}
Now, we uncover the energy-accuracy trade-off associated with each design
parameter by studying its impact on the training time and prediction accuracy
of the model. To begin, we focus on two memory parameters: EM and SB.

We explore EM and SB for different pairs of sizes and show the accuracy and
energy usage in~\autoref{fig:memory_params_tiny} for CIFAR100 (a/b) and
Tiny-ImageNet (c/d) using ER. BiC exhibits similar trends, so we omit it for
brevity. Each accuracy figure of (a) and (c) includes a set of lines, with
each line representing the changes in accuracy for a particular EM size over
different SB sizes on the x-axis\footnote{Note that the task size of CIFAR100
and Tiny-ImageNet is 5K and 10K. If the SB is smaller than the task size,
some remaining samples are not stored in SB but are still kept in storage and
retrieved later for replay.}. The energy property in (b) and (d) is easy to
depict -- More energy is consumed for larger memory combining more EM and SB samples. 
Then, the question remains whether using larger memory is
cost-effective.

For the same EM size, increasing the SB size does not necessarily improve
accuracy.
In many cases, the highest accuracy is achieved when the SB size is smaller
than the maximum value on the x-axis. This phenomenon significantly harms
energy-sensitive edge devices because a larger SB directly leads to longer
training time and higher energy spending. The red symbol
\textcolor{red}{$\star$} on each accuracy graph indicates the SB size that
produces the best accuracy for the same-sized EM. The optimal SB size often
differs.

We observe a similar trend when adjusting the EM size for the same-sized SB,
i.e., the same x-axis value. Specifically, for both CIFAR100 and
Tiny-ImageNet, we see no noticeable improvement in accuracy when the EM size
increases from 18K to 30K. A bigger EM also leads to increased data swapping
and I/O operations due to more EM samples drawn for training. However, the
I/O power consumption is insignificant relative to the GPU power, so an
enlarged data-swapping activity does not contribute to higher energy
consumption.

\begin{figure}[!t]
\centering
\includegraphics[width=1\linewidth]{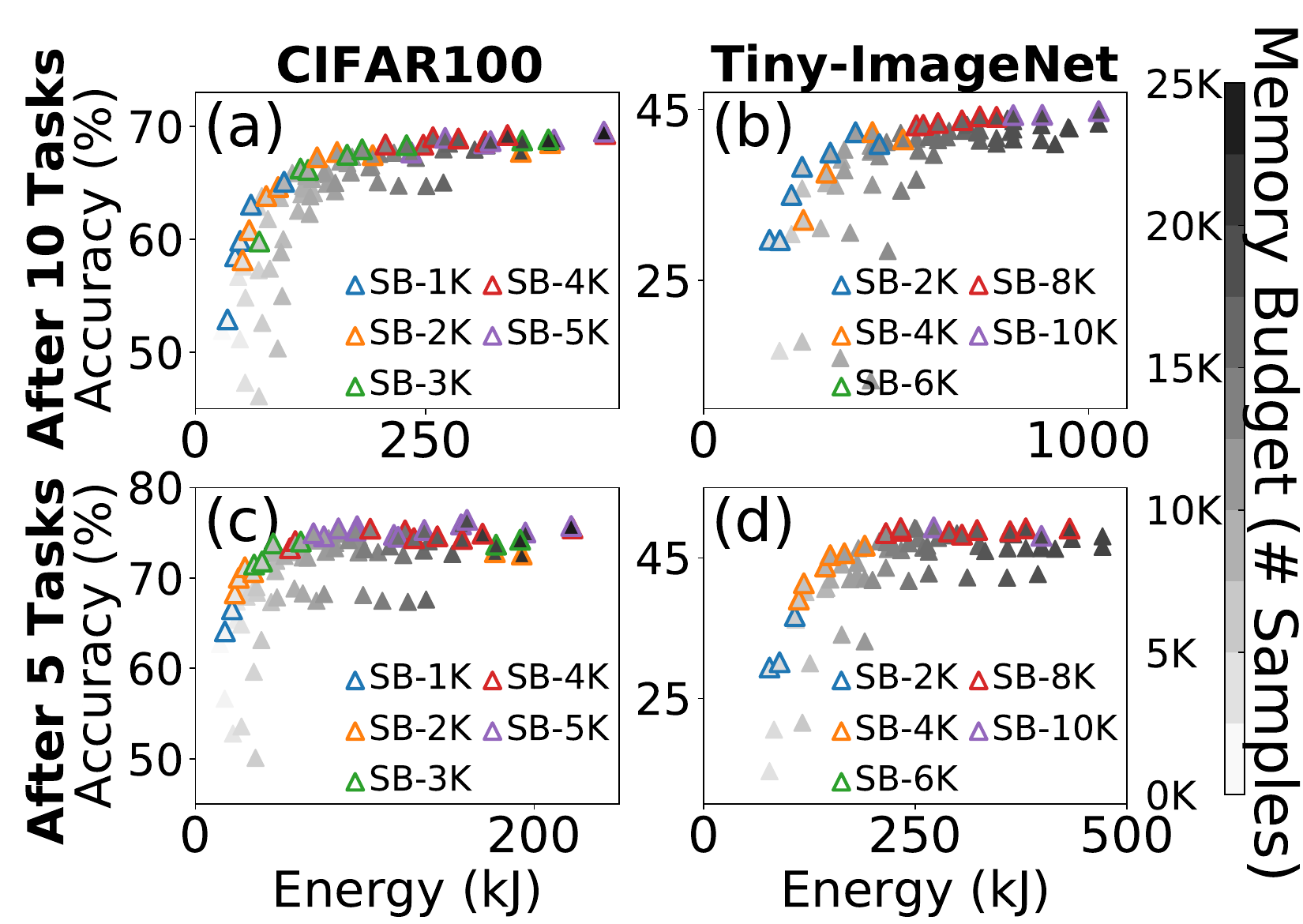}
\vspace{-0.2in}
\caption{Energy-accuracy trade-offs over varying memory budgets shared by EM and SB.}
\vspace{-0.2in}
\label{fig:config_to_accuracy_and_memory_budget}
\end{figure}

\parlabel{EM and SB as One.}
From \autoref{fig:memory_params_tiny}, it is evident that effectively
allocating a given memory budget across EM and SB is a challenging task. To
further demonstrate this, we show the impact of different EM and SB size
pairs on energy and accuracy while varying the memory budget
in~\autoref{fig:config_to_accuracy_and_memory_budget}(a) and (b). Each data
point is represented by various shades of gray based on the memory budget,
with darker shades indicating higher memory budgets.
We also highlight the color of the SB size that produces the best accuracy
out of all possible size pairs that split the same memory budget. We make
three observations. (1) Bigger memory budget still does not guarantee higher
accuracy. (2) The best SB size differs across memory budgets. (3) Blindly
committing full capacity
\textcolor[rgb]{0.59, 0.44, 0.84}{$\triangle$} to SB to emphasize ``learning
a new task'' over ``remembering old tasks'' is not always necessary.

The current approach to allocating memory \emph{prioritizes} SB over EM to
enable all new samples to be used for training~\cite{gdumb, icarl, bic}. This
approach is reasonable for single-level EMs in memory, where only a small subset of
samples from an old task can be stored 
and participate in replaying.
However, HEM maintains a much larger number of old samples in storage,
allowing for more effective replay. This is presumably the main reason behind
the third observation. We observe no static memory pair that aces across NN
models, datasets, and problem sizes, which renders static or history-based
memory-sizing strategy ineffective. For
example,~\autoref{fig:config_to_accuracy_and_memory_budget}(c) and (d) show
accuracies
right after the training of five tasks. The best accuracy for each memory
budget is vastly different compared
to~\autoref{fig:config_to_accuracy_and_memory_budget}(a) and (b). These
findings are summarized in the second row of~\autoref{tab:insights}.

\begin{figure}[!t]
\centering
    \includegraphics[width=\linewidth]{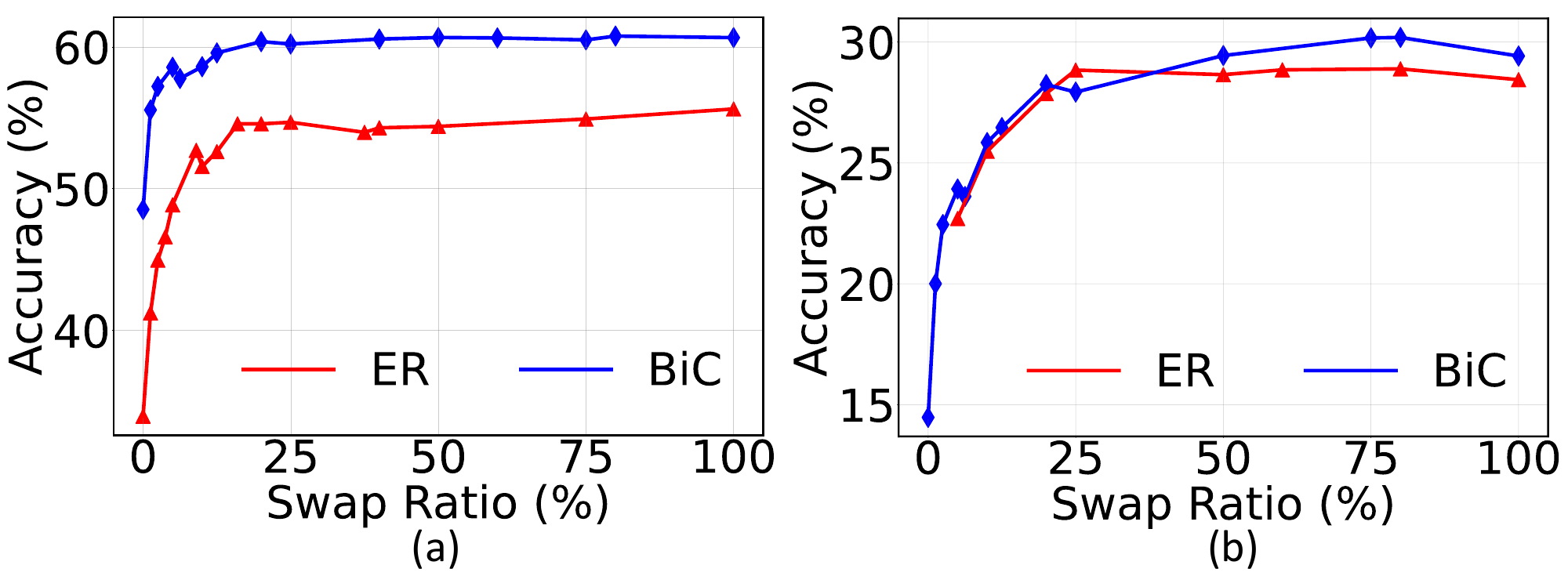}
    \vspace{-0.2in}
\caption{Accuracy over varying swap ratios for CIFAR100 (a) and Tiny-ImageNet (b).}
\vspace{-0.2in}
\label{fig:swap_grids}
\end{figure}

\parlabel{I/O Parameter: Swap Ratio.}
\autoref{fig:swap_grids}(a) and (b) show the impact of the swap ratio on
model accuracy for ER and BiC with an EM size of 2K, using CIFAR100 and
Tiny-ImageNet, respectively. We see that even a tiny swap ratio can
remarkably improve accuracy compared to traditional non-hierarchical EM
setups, which correspond to a swap ratio of zero. Overall, the accuracy
increases sharply from 0\% to 20\% and then plateaus.
This trend is fairly consistent across EM sizes.

Notice that HEM's I/O operations are asynchronous and incur virtually zero
delays to training time~\cite{carm}. Moreover, the required I/O bandwidth for
full swapping, where every EM sample is swapped every epoch, is only 3.8MB/s
for CIFAR100 and 16.8MB/s for Tiny-ImageNet, which are well below the I/O
bandwidth provided by removable storage devices like MicroSD cards. Faster
mobile GPUs that can train models faster may put more strain on I/O traffic.
However, due to the insignificant impact of I/O on system dynamic power (\ie,
0.1W for 100\% swap ratio), faster mobile GPUs are unlikely to exacerbate
I/O power in the near future. So, improving model accuracy with a higher swap
ratio is nearly cost-free.
This claim is also factored in the first row of~\autoref{tab:insights}.

\begin{figure}[!t]
\centering
\includegraphics[width=\linewidth]{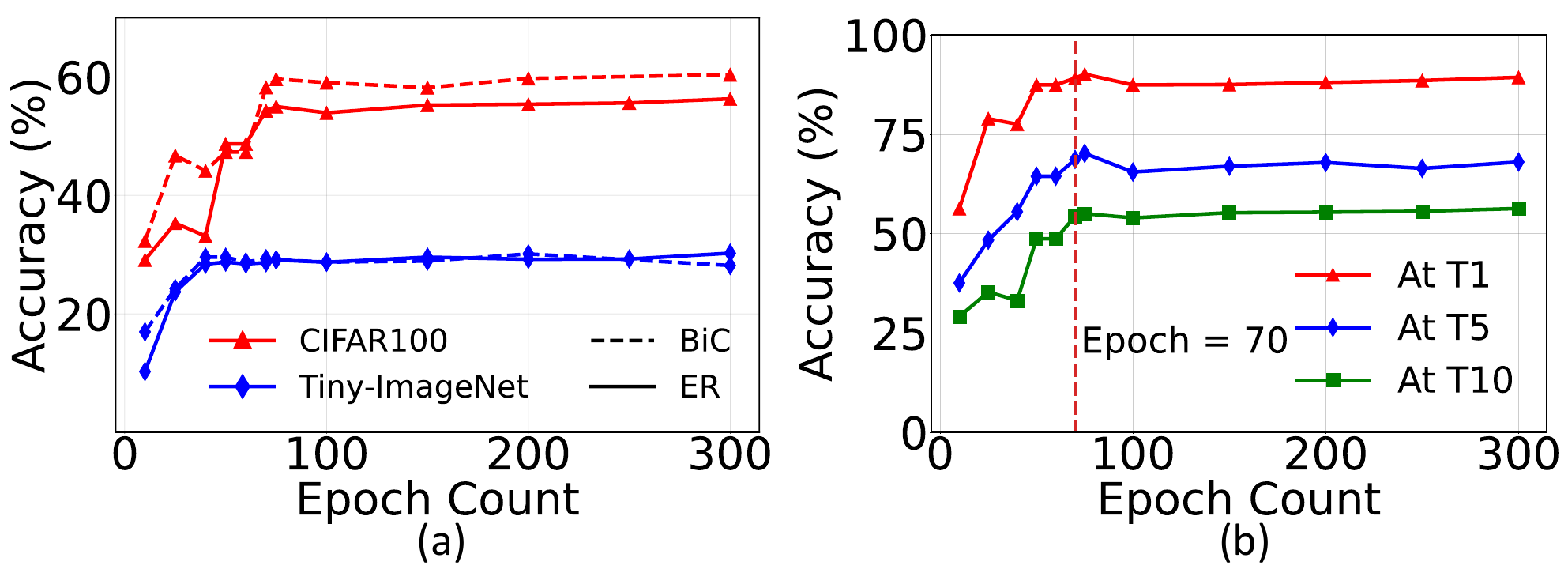}
\vspace{-0.2in}
\caption{Accuracy over varying epoch counts.}
\vspace{-0.2in}
\label{fig:accuracy_epochs}
\end{figure}

\parlabel{GPU Parameter: Epoch Count.}
The system consumes energy proportional to the epoch count. Typically, for
both ER and BiC, accuracy reaches a plateau after a certain number of epochs,
as shown in~\autoref{fig:accuracy_epochs}(a). This indicates that a statically
configured epoch count is feasible. Then, a key question is whether we can
terminate training for each task earlier to save energy. To see the
possibility, we explore the accuracy observed after training the first,
second, and last tasks using different epoch counts. For brevity, we present
the results for ER on CIFAR100 in~\autoref{fig:accuracy_epochs}(b). The
vertical line indicates the epoch count of 70, a hyperparameter suggested by
prior work~\cite{carm}. The number of epochs required to reach convergence
remains almost identical at around epoch 70. Reducing the number of epochs
for intermediate tasks has negatively affected subsequent tasks. Therefore,
we consider using the value proven to work well without any adaptation. This
is the last insight we describe in the third row
of~\autoref{tab:insights}.

\section{\runtime: System Runtime}
\label{sec:system_runtime}

We develop a system runtime, \runtime, to showcase how our insights for HEM
can be leveraged. \runtime focuses on optimizing the dynamic parameters
defined in the second rightmost column in~\autoref{tab:parameters}.
Specifically, \runtime accomplishes two critical missions for cost-effective
HEM while adhering to device-specific resource constraints:

\emph{1. Selecting data swapping strategy.} The swap ratio is a
capacity-bound parameter that has a relatively minor impact on overall energy
expenditure. Thus, \runtime sets a swap ratio that attempts to fully utilize
the available I/O bandwidth.
To achieve the goal, \runtime continuously monitors system-wide I/O usage to
refine the data swapping strategy.

\emph{2. Deciding stream buffer and EM sizes.}
The optimal combination of SB and EM, which exhibits high cost-effectiveness,
differs across memory budgets and tasks.
Thus, \runtime dynamically sizes both SB and EM in the presence of
consecutive new tasks within the memory budget. As discussed, using more
memory incurs higher energy costs but does not necessarily guarantee higher
accuracy. This makes accurately sizing SB and EM even more crucial.

\begin{figure}[!t]
    \centering
    \includegraphics[width=\linewidth]{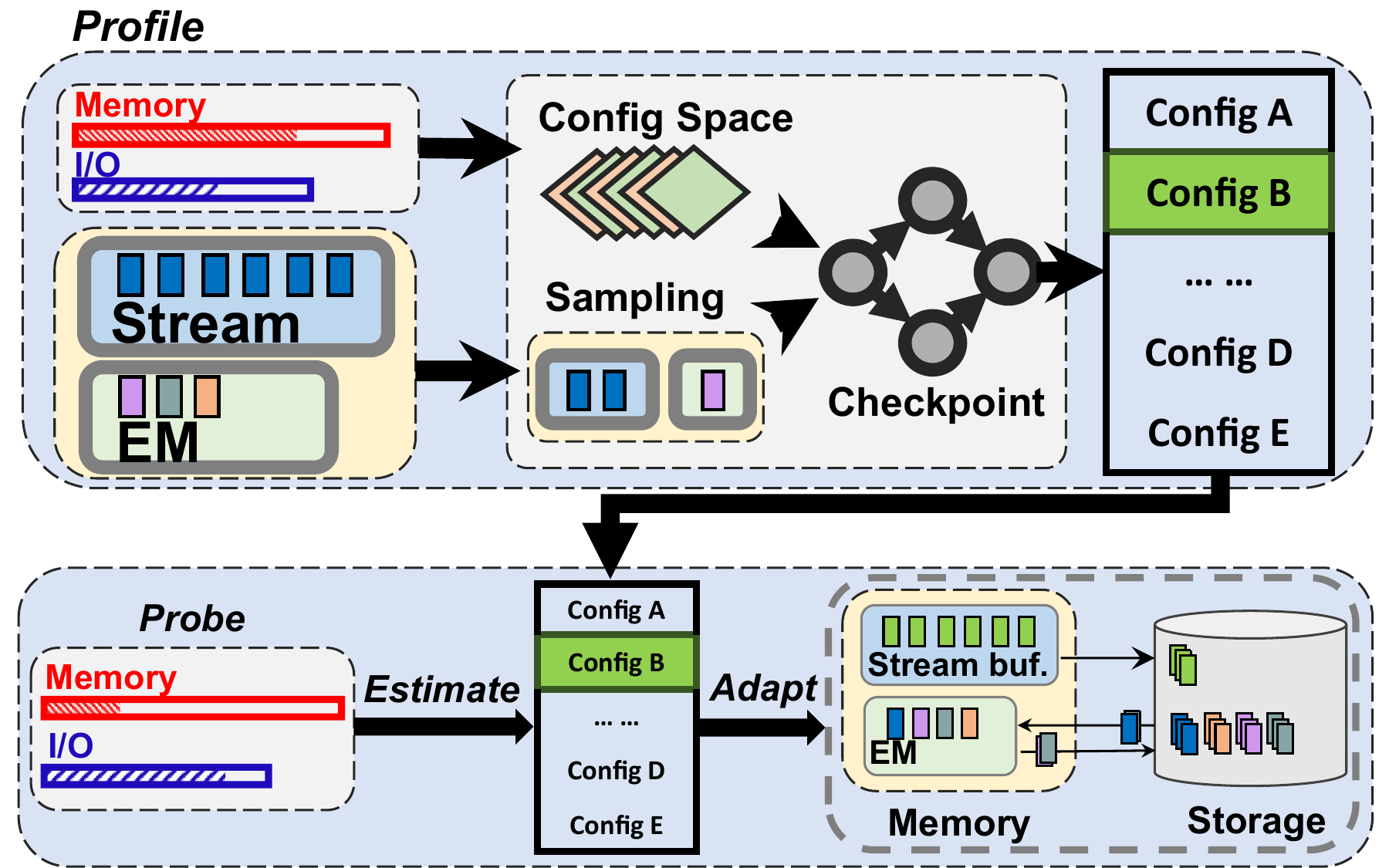}
    \vspace{-0.1in}
    \caption{Miro system runtime architecture.}
    \vspace{-0.2in}
    \label{fig:runtime_arch}
\end{figure}

\subsection{Design Overview}
\label{sec:design_overview}

Although the two missions outlined above are seemingly independent, we strive
to integrate them into a unified workflow that operates through a
well-defined state machine. This approach allows us to optimize individual
parameters without disrupting other aspects of runtime execution. Here, we
describe \runtime's operational phases as shown in~\autoref{fig:runtime_arch}:

\begin{myitemize}

    \item \textbf{Profile:} When a new task is ready, \runtime conducts
    profiling by focusing on those parameters constrained by cost-efficiency, i.e.,
    SB and EM. This profiling involves searching for
    various SB-EM size pairs based on possible memory
    allocations within a memory budget. For each size pair, \runtime records
    an estimate of accuracy and energy usage associated with using it for the
    new and old tasks.

    \item \textbf{Probe:} At the end of every epoch, \runtime probes system
    resource states to see whether any parameters need reconfiguration.
    \runtime focuses primarily on I/O and memory states for data swapping and
    SB-EM sizing, respectively, and enters \texttt{Estimate} phase
    if a state change is identified: e.g., I/O becomes congested or the memory
    budget is updated.

    \item \textbf{Estimate:} \runtime computes new parameter values for the
    state change occurring at the target resource. It computes
    a new swap ratio for I/O congestion or a new SB-EM size for a new memory
    budget. Then, \runtime enters \texttt{Adapt} phase.

    \item \textbf{Adapt:} \runtime refines the target parameters
    with the values derived from \texttt{Estimate} phase.
    Once this refinement is complete, \runtime re-enters \texttt{Probe} phase.

\end{myitemize}

\runtime carries out profiling of SB and EM
at every task insertion, requiring it to be done at low overhead. While there
are other approaches to avoid repetitive profiling, such as relying on static
configurations or historical data, we have opted \emph{not} to pursue them
for the following reasons:

\parlabel{Why Not Using Static Configurations?}
A static configuration often produces the best outcome when recurring
training jobs in an offline setup start from the same initial
condition~\cite{zeus}, i.e., a fresh model with no observed data. However, in
continual learning, each task begins training from a distinct condition due
to the accumulation of model updates over time and data drift occurring among
tasks. Thus, a configuration that yields optimal results for the current task
may not produce the same outcome for the next tasks.

\parlabel{Why Not Using Configurations from History?}
An incoming task $T_i$ may resemble an old task $T_j$,
thereby tempting us to reuse the same configuration that was used
for $T_j$. However, despite the similarity between the stream samples of
$T_i$ and $T_j$, the samples in EM may differ significantly. In other words,
the knowledge that the model must retain upon the availability of $T_i$ as
compared to $T_j$ may be markedly different. So, we cannot guarantee that a
previously successful configuration will be effective for similar tasks in
the future.

\subsection{Data Swapping Strategy}

The strategy for data swapping in \runtime is designed with three properties
in mind:

\begin{densedesc}

\item[\textbf{P1)}] I/O energy consumption is insignificant.

\item[\textbf{P2)}] Increasing the swap ratio provides benefits across a
    broad range, with a knee point appearing at a relatively low swap ratio
    (15--20\%) in the ratio-accuracy curve, as demonstrated
    in~\autoref{fig:swap_grids}.

\item[\textbf{P3)}] Other programs running on the device can abruptly
    compete for I/O resources.
    But, under normal circumstances, HW typically allows training jobs to
    leverage ample bandwidth for full-fledged data swapping.

\end{densedesc}

\parlabel{Tuning Swap Ratio.}
\textsf{P1} allows us to focus on a single knob, I/O usage. When I/O is
idling, we increase the ratio \emph{incrementally} by a small discretized
value (default 10\%), taking advantage of the benefits highlighted in
\textsf{P2}. Conversely, when I/O is congested, we decrease the ratio by a
large quantity (default half) to \emph{quickly} alleviate the I/O contention described in
\textsf{P3}. Despite the ratio that decreases rapidly, we can
maintain an effective swap ratio even after a few times of congestion
handling, i.e., $100\% \rightarrow 50\% \rightarrow 25\%$, without reaching
the knee point. The techniques we use to adjust the swap ratio based on I/O
contention or idleness are skin to those employed in TCP congestion
control~\cite{tcp_reno, tcp_cubic}. All this process is performed in
\texttt{Estimate} phase.

\parlabel{Obtaining I/O State.}
During \texttt{Probe} phase, \runtime performs system I/O queue monitoring to
determine the state of HEM, which can be classified as one of three states:
\emph{Congested} when the I/O queue experiences back-pressure;
\emph{Idle} when the I/O queue remains empty for longer than a predefined
duration of time; and \emph{Stable} when the I/O queue is neither congested
nor idle. If I/O is idle or congested,
\runtime forwards the observed state to \texttt{Estimate} phase, which in
turn triggers the adaptation of the swapping ratio to react to the state.

\parlabel{Reducing Swap Ratio.}
\runtime reduces the swap ratio by first increasing the swap interval. We
found that updating the interval value is beneficial for swap ratios ranging
from 100\% to 20\%, which corresponds to a value range of 1 to 5 epochs.
However, for swap ratios below 20\%, we maintain the interval fixed at 5
epochs and adjust the swap percentage adequately to achieve the desired
target swap ratio.

\subsection{Stream Buffer and EM Sizes}

Now, we explain how \runtime configures two memory parameters, SB and EM,
that significantly influence both accuracy and energy efficiency. The sum of
their sizes should not exceed the memory budget, which we assume to be
user-specified by system admins.

\subsubsection{Memory Parameter Optimization Problem}

\runtime must reduce the strain on small battery-powered devices that require
high energy efficiency while delivering high-accuracy models to satisfy user
experiences. Between energy efficiency and accuracy, which are often at odds
with each other, we assume accuracy takes precedence over energy efficiency.
Therefore, our ultimate objective is to identify an optimal configuration for
SB and EM sizes, or \emph{conf} for brevity, that conserves energy while
maintaining high accuracy. Since the definition of high accuracy can be
subjective, \runtime provides users with an implicit method called
\emph{cutline} to aid in determining its appropriate magnitude.

\subsubsection{Design Insights through an Example}

\begin{figure}[!t]
\centering
\includegraphics[width=\linewidth]{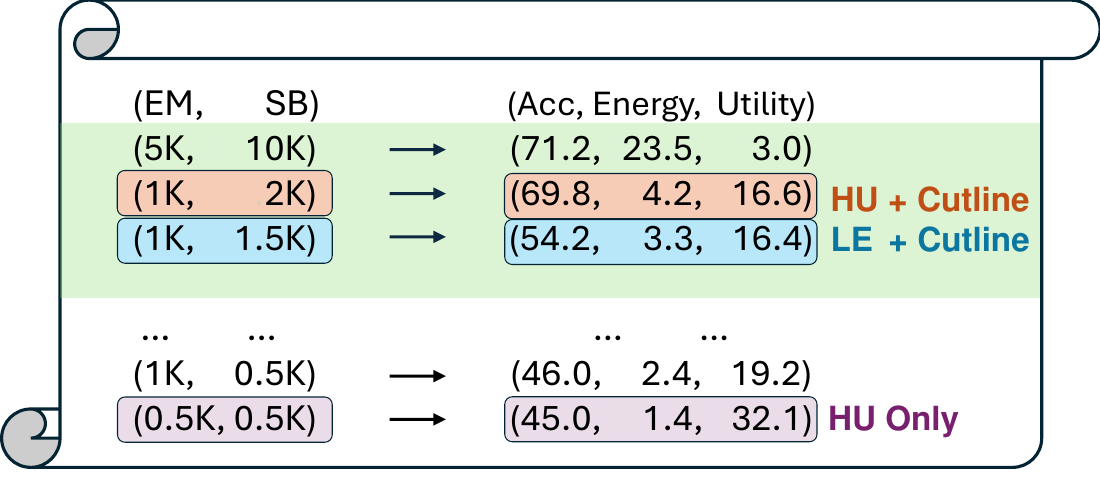}
\vspace{-0.1in}
\caption{An illustrative example of how our method works. HU: highest utility. LE: lowest energy.}
\vspace{-0.2in}
\label{fig:illustrative_example}
\end{figure}

To compare our algorithms with existing memory-optimizing algorithms and explain the rationale behind our choice,
we use a simplified
example borrowed from real profiling as shown
in~\autoref{fig:illustrative_example}. This example includes a list of confs
along with their respective accuracy and energy consumption 
associated
with each conf. Later in this section, we will discuss our
profiling method for acquiring this information at low costs.

\parlabel{Why Not Using a Single Metric?}
Consider methods that select a conf using a single metric, either energy
efficiency or accuracy. The most energy-efficient conf
in~\autoref{fig:illustrative_example} is (0.5K, 0.5K) for (SB, EM). In this
case, the achieved accuracy is far below what we believe as good accuracy
shown in the list. Similarly, the conf that delivers the highest accuracy is
(5K, 10K) that can store up to 15K samples in total, making it the
highest energy-consuming option. However, if we examine confs with similar
accuracy levels, e.g., within 2\%, we can find a more energy-efficient option
that requires only 3K samples, the conf (1K, 2K). This conf requires around
20\% of GPU energy compared to the 15K-sample conf. We commonly
encounter similar problems when dealing with many lists of memory confs using
a single metric.

\parlabel{Simply Combining Two Metrics is Enough?}
To account for both metrics simultaneously, we propose a new metric called
\emph{utility}, which represents the accuracy gained per unit of the energy
spent\footnote{We tested various utility score designs, including those based
on curve-fitting and stochastic approximation. However, they either proved to
be computationally expensive or did not outperform our current design.}:
\vspace{-0.2cm}
\begin{equation}
  \text{Utility}=\frac{\text{Accuracy\ Gain}}{\text{Energy\ Usage}}
\end{equation}
\noindent Therefore, the conf with the highest utility is the one that spends
energy most efficiently to improve accuracy.

\autoref{fig:illustrative_example} shows that the conf (0.5K, 0.5K) has the
highest utility. 
To maintain good accuracy in this round of 
training, however, one might choose the
conf (1K, 1.5K) or the conf (1K, 2K) instead
as both are considered
energy-efficient among all candidates.
This highlights that merely considering utility, although both energy
efficiency and accuracy are factored in, falls short of meeting our
fundamental requirement of achieving ``high accuracy''. To reach our
objective, we need to first exclude unpromising confs attaining low accuracy.

\subsubsection{Our Method.}

We introduce \emph{cutline}, a straightforward yet powerful technique that
enables us to narrow down high-accuracy confs by
their accuracy ranks. For instance, if we apply a cutline of 20\% to our
example in~\autoref{fig:illustrative_example}, we can obtain the candidate
subset that comprises 20\% of the confs (colored region) with accuracy higher than the other
80\% that the cutline has filtered out. On top of this, if we search for the
conf with the best utility, we obtain the conf (1K, 2K), which is a valid
choice. Interestingly, we can even apply a single metric ``energy
efficiency'' after the cutline and obtain a good conf (1K, 1.5K). This is
because the cutline already sifts out candidates that satisfy our
high-accuracy criterion.

Using a proper cutline is undoubtedly crucial, involving a
trade-off. A tighter cutline will leave fewer confs with higher accuracy,
thereby increasing the likelihood of selecting a conf that consumes more
energy. Based on empirical studies, we found that a cutline range of 20--50\%
works well in practice.

Nonetheless, in scenarios where a very slight margin from the maximum
possible accuracy is permitted, \runtime can 
dynamically determine the
cutline to identify a small group of confs with exceptionally high accuracy.
It is important to note that larger memory usage does not guarantee higher
accuracy, and achieving the highest accuracies over task insertions requires
different memory allocations between SB and EM. \runtime addresses
these challenges, regardless of the cutline value in use.

\subsubsection{Profiling at Low Overhead}

The goal of profiling is to build a fresh list of confs to examine for every
task insertion. Profiling should be lightweight so that its energy expenses
do not outweigh the benefits gained from executing a cost-effective conf
selected by \runtime. Na\"{i}ve profiling entails model training on many
different SB and EM sizes for many epochs and samples, which leads to
prohibitively high energy costs. As such, we strive to minimize profiling
overheads from two angles: (\textsf{A1}) number of confs and (\textsf{A2})
single-conf evaluation time. We follow several popular guidelines:

\begin{densedesc}

\item[\textbf{A1)}] Avoid exhaustive profiling that covers all size
    variations. Profile a small subset of confs.

\item[\textbf{A2.1)}] Do not use the entire training data that includes
    all stream buffer and EM samples. Use a subset of the data.

\item[\textbf{A2.2)}] Do not go through all epochs. Perform training for a
    small number of epochs and infer the accuracy that could be obtained
    if there were many more epochs.

\end{densedesc}

While there are common conventions taken to tackle these issues, they do not
work very well on our problem or should be used with caution:

\parlabel{Reducing \# Confs (A1).}
Several algorithms or heuristics~\cite{ekya} exist to narrow down the search
space. The idea behind 
is to identify unpromising confs in history and
exclude them from the search range.
However, in CL setups,
confs from the past may not provide much insight into their significance in
the future due to similar reasons explained
in~\autoref{sec:design_overview}.
Therefore, we opt for a safer approach of reducing the number of confs by
uniformly sampling instances from the space.

\parlabel{Reducing \# Training Samples (A2.1).}
We rely on random sampling to reduce training samples, as it is well-known
for preserving the distribution of original data in CL setups~\cite{ekya,
carm}. On top of it, we meticulously verify if the sampling rate in use
covers samples across all classes. Otherwise, the classes not represented in
the samples will report zero accuracies, causing misleading results.

\parlabel{Reducing \# Epochs (A2.2).}
We obtain the accuracy of each conf by running a few epochs and use the
observed accuracy without extrapolating to the final accuracy of the model.
In general, confs that achieve higher accuracy when processed until
convergence tend to have a steeper accuracy increase during an earlier
training phase. So, the accuracy observed through several training epochs is
just informative enough for us to apply a cutline, rank confs, and measure
utilities.

\begin{figure}
    \centering
    \includegraphics[width=\linewidth]{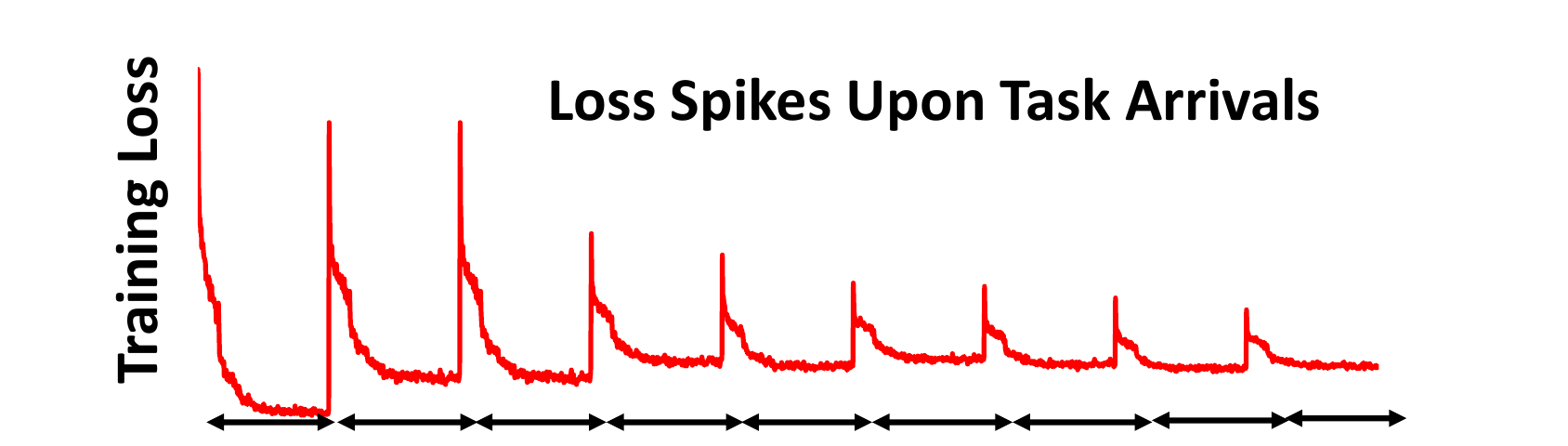}
    \vspace{-0.2in}
    \caption{Constant spikes and noises in training loss when training new tasks.}
    \vspace{-0.2in}
    \label{fig:loss_spikes}
\end{figure}

However, one major challenge we face is deciding an appropriate time to
collect the accuracy data. We have observed that during the earlier epochs,
the model tends to exhibit spiky and noisy patterns in
training loss that do not reveal any
useful information about the accuracy trend across new and old tasks. 
\autoref{fig:loss_spikes} illustrates this phenomenon. 
It is only after the first few epochs that the model
begins to smoothly improve loss values for both new and old tasks in harmony.
Therefore, we take a model checkpoint from a reference conf that has reached
a stable point and use it to profile all confs. More
importantly, this approach saves on profiling costs tremendously because it
eliminates the need to navigate through the earlier phase for each conf.

\subsubsection{Tying All Together in \runtime Workflow}

During \texttt{Profile} phase, \runtime selects a set of random confs
(default 14 confs) that can fit in the memory budget. It then creates a
checkpoint model using a baseline conf borrowed from prior work \cite{icarl,
carm}. Using the checkpoint, each conf runs for a few epochs (default 5
epochs) on a small subset (default 5\%) of training data. After all confs are
evaluated, \runtime applies a cutline and selects the conf with the highest
utility. At the end of every epoch, \runtime invokes \texttt{Probe} phase to
detect if the memory budget is updated. Since the conf selected by our method
does not typically harness the full memory budget, SB and EM sizes do not
need to be reorganized every time. Only when the current memory usage exceeds
the new budget, \texttt{Estimate} phase selects the conf with the highest
utility from previously profiled results (without re-profiling) while
excluding those confs that the new budget cannot serve. If the budget
increases, \runtime performs profiling in the next task to avoid excessive
profiling. Finally, \texttt{Adapt} phase reconfigures SB and EM based on
their new parameter values.

\section{Implementation}
\label{sec:implementation}

\runtime is open source at https://github.com/omnia-unist/Miro.
Built atop CarM~\cite{carm_code}, \runtime extends three key components:

\parlabel{Episodic Memory.}
\runtime implements EM through shared memory using the Python
Multiprocessing module~\cite{sharedmem} to enable direct reads and writes on
a shared address space. These accesses are not serialized nor
synchronized for high speed, so they can interfere with each other
in the same memory location. We ignore this for two reasons: (1) rare event
and (2) no visible effect on the statistical model efficiency.

\parlabel{Monitoring.}
Instead of directly consulting with the OS on I/O status, \runtime probes the
completion rate of recent data swapping as a proxy of I/O congestion or
idleness. This approach reduces monitoring costs. When the swap ratio is
reconfigured, \runtime informs the swap worker in CarM to change the swapping
strategy. The swap worker is also notified of any newly configured EM
addresses and sizes.

\parlabel{Profiling.}
\runtime constructs a conf search space by taking into account the new task
size and the current memory budget. It then launches mini-training tasks to
quickly obtain accuracy and energy estimates for each candidate conf. To
avoid frequent SB or EM buffer resizing during profiling, \runtime masks the
buffers when the target size is smaller than the current buffer size. This is
the common case with the sub-sampled buffers used in profiling.

\section{Evaluation}
\label{sec:evaluation}

\begin{figure*}[!t]
\centering
    \includegraphics[width=0.95\linewidth]{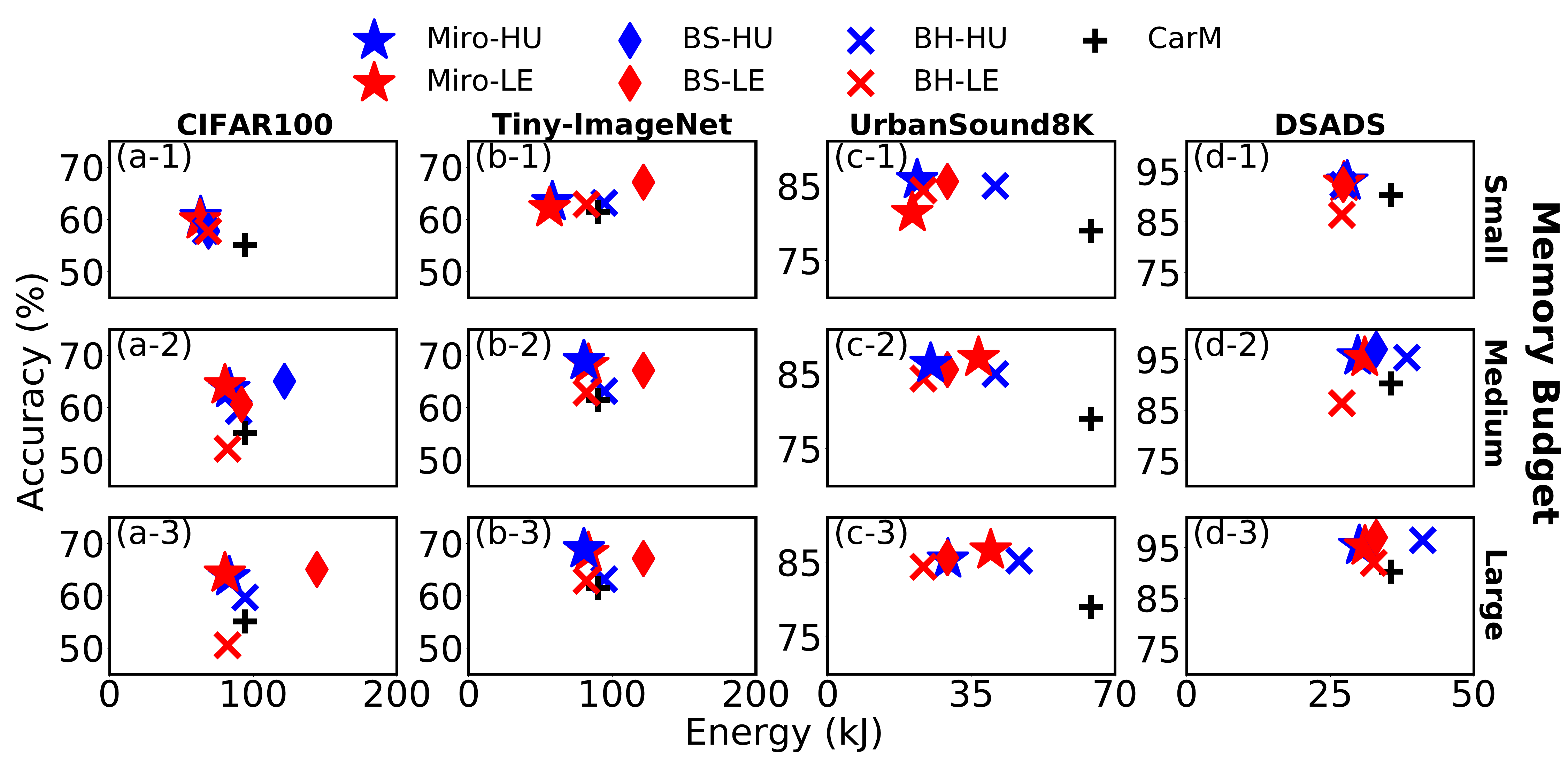}
    \vspace{-0.15in}
    \caption{Energy-accuracy trade-offs over competing static methods for a combination of different datasets (a-d) and memory budgets (1-3). For example, the subgraph (a-1) compares the methods using CIFAR100 on a small memory budget. The memory budgets are 10K, 25K, and 50K for CIFAR100 and Tiny-ImageNet and 1K, 2K, and 2.5K for UrbanSound8K and DSADS.}
\label{fig:baselines}
\vspace{-0.15in}
\end{figure*}

\begin{table}[!t]
    \centering
    \resizebox{0.48 \textwidth}{!}{%
    \begin{tabular}{|c|c|c|c|c|c|}
    \hline
        \textbf{Dataset} & \textbf{Classes} & \textbf{Tasks} & \textbf{Samples} & \textbf{Size} & \textbf{Task Domain } \\ \hline
        CIFAR100 & 100 & 10 & 50k & 197MB & Image Classif.  \\ \hline
        Tiny-ImageNet & 200 & 10/20 & 100k & 1.1GB & Image Classif.  \\ \hline
        UrbanSound8K & 10 & 10 & 8732 & 184MB & Audio Classif.  \\ \hline
        DSADS & 19 & 10 & 9120 & 163MB & HAR  \\ \hline
        ImageNet1k & 1000 & 50 & 1.28m & 146GB & Image Classif.  \\ \hline
    \end{tabular}}

    \caption{Datasets used in \S \ref{sec:characterization} and \S \ref{sec:evaluation}. We cover three task domains: Image Classification, Audio Classification, and Human Activity Recognition (HAR).}
    \label{tab:datasets}
    \vspace{-0.2in}
\end{table}

Our experimental setup builds upon the methodology described
in~\autoref{sec:methodology}, with several additional benchmarks outlined
in~\autoref{tab:datasets}. (1) \textbf{Audio Classification}: We use
ResNet-18 on the UrbanSound8k dataset~\cite{urbansound}, which contains 8,732
urban sound files categorized into 10 classes. The sound files are
transformed into log-mel spectrograms following the approach
in~\cite{Salamon2016DeepCN}. The dataset is split into 10 single-class tasks.
(2) \textbf{Human Activity Recognition}: We use ResNet-18 on the Daily and
Sports Activities dataset~\cite{dsads}, which contains 9,120 motion sensor
data samples covering 19 daily and sports activities. The dataset is split
into 10 tasks, with tasks 1 to 9 having 2 classes and task 10 having 1 class.
(3) \textbf{Large-scale Image Classification}: We organize the ImageNet1k
dataset into 50 tasks, denoted as ImageNet1k-50Tasks, with each task
comprising 20 classes.

In addition, we revise the Tiny-ImageNet to test 20 tasks for a longer
sequence, with each task comprising 10 classes, while the CIFAR100 remains
with 10 tasks~\footnote{For ImageNet1k-50Tasks, we perform extrapolation
to a larger-memory device by utilizing a server with more memory capacity to
faithfully evaluate this demanding dataset, assuming \runtime operates on a
higher-end edge device with similar energy properties.}.
We use ER as the reference model since HEM makes it optional to use
sophisticated algorithmic optimizations. All experiments are conducted on
Jetson Xavier NX.
By default, our system consistently consumes 4.7--6GB of memory: 2GB for system
bootup and 2.7--4GB for running CL methods in PyTorch on unified memory
shared by the CPU and GPU.

\parlabel{Competing Methods.}
Since \carm is the only existing method that exploits memory hierarchy for
CL, we have additional baselines and cover strategies that incorporate better
static confs, historical data, and heuristics.
(1) \underline{\carm} employs static confs as suggested by~\cite{carm}. (2)
\underline{\bsfull} explores all possible static confs and finds the conf
that offers the best cost-effectiveness, guided by a cutline and the highest
utility or lowest energy. This is intended for us to see the (almost) best
possible outcome for using static confs. (3) \underline{\bhfull} selects the
best intermediate conf identified by \bsfull after completing half of the
tasks and uses the conf for future tasks. (4) \underline{\heufull} treats new
and old tasks equally and assigns memory to SB \vs EM proportional to the
number of tasks placed in each component. The motivation behind this is that
HEM can replay old data very effectively via data swapping. For a fair comparison, we assume
all methods perform full-fledged data swapping unless otherwise specified.

We adopt a step size of 0.5K samples for memory sizing, which is sufficiently
fine-grained. In reality, \bsfull and \bhfull require exhaustive exploration
of numerous static confs, but we exclude the exploration cost for these
methods to focus on showing their potential best-case efficacy. However, we
include the profiling overhead in \runtime to avoid over-emphasizing our
effectiveness. We will show that despite this overhead, \runtime outperforms
the other methods.

\subsection{Small to Medium-Scale Training}

We begin our experiments by evaluating all competing methods on 
the first four
datasets from~\autoref{tab:datasets} in two steps.
First, we compare \runtime with \carm, \bsfull (\bs), and \bhfull (\bh),
which represent static strategies. Second, we compare \runtime with \heufull
to focus on
dynamic strategies. A na\"ive \heufull (\heu) always uses up the entire
memory budget, so it clearly hurts energy efficiency. Therefore, we
allow \heufull to consume a portion of the memory budget, enabling it to
trade off accuracy for energy saving.
\parlabel{Comparing to Static Methods.}
To perform a faithful comparison of the competing methods, we vary memory
budgets over the datasets and draw energy-accuracy trade-off graphs
in~\autoref{fig:baselines}.
For \runtime, \bsfull, and \bhfull, which employ a cutline, we set it to 50\%
and show the conf choices from both the highest utility (\textsf{HU}) and
lowest energy (\textsf{LE}) perspectives. In each graph, the x-axis
represents energy usage (lower is better),
and the y-axis represents accuracy (higher is better).
Cost-effective methods will choose confs that 
achieve high accuracy while saving energy,
appearing closer to the upper-left corner of the
graph.
Overall, \runtime
consistently outperforms or is on par with other methods, demonstrating
superior cost-effectiveness.

\begin{figure}[!t]
\centering
    \includegraphics[width=0.95\linewidth]{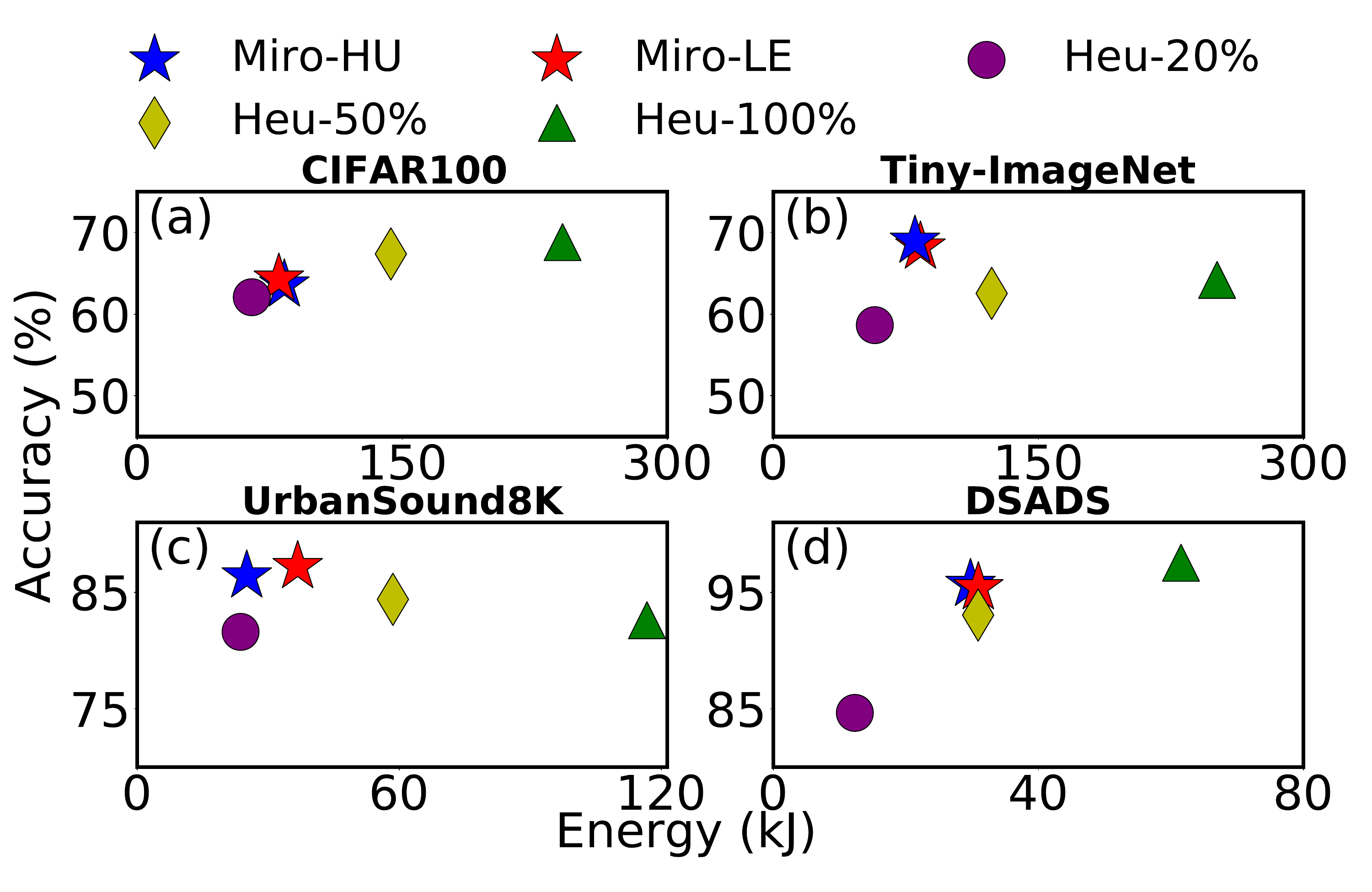}
    \vspace{-0.1in}
    \caption{Energy-accuracy trade-offs between \runtime and \heufull over different datasets at the medium memory budget used in~\autoref{fig:baselines}.}
\label{fig:heuristic}
\vspace{-0.2in}
\end{figure}

It is worth comparing \runtime with \carm, the state-of-the-art HEM design
based on static parameters, in detail. Compared to \carm, \runtime uses
23--66\% less memory while achieving 1.35--9.19\% higher accuracy across
benchmarks and memory budgets. 
Overall, \runtime shows higher effectiveness when larger memory budgets are available.
This is because our
runtime profiler can cover a wider spectrum of EM and SB sizes and determine
more effective confs that can be found in larger memory budgets.

\runtime, in particular \runtime-\textsf{HU}, tends to make better choices
than \bsfull and \bhfull, even without including their prohibitive
exploration costs in the graphs. Occasionally, they provide confs with
slightly better accuracies but at significantly higher energy usage. \bhfull
is less effective than \bsfull even though they both require exploration to
find out the best static conf. This indicates that previously successful
confs may not be effective for future tasks. To summarize, \runtime achieves
its goal of striking a good energy-accuracy trade-off across memory budgets.

\parlabel{Comparing to Dynamic Methods.}
For \heufull, we vary the amount of in-use memory in order to avoid always
utilizing a given memory budget, as shown in~\autoref{fig:heuristic}. When
\heufull fully utilizes the given memory (\heu-\textsf{100\%}), it provides an accuracy improvement
of only 4.45\% for CIFAR100 and 2.24\% for DSADS, while consuming three times
more memory on average than \runtime. This trade-off is not considered
worthwhile. For UrbanSound8K and Tiny-ImageNet, \runtime outperforms \heufull
in accuracy by 4.6\% and 4.04\%, respectively.
When we limit \heufull to use 20\% (\heu-\textsf{20\%}) or 50\%
(\heu-\textsf{20\%}) of the given memory budget, it
occasionally achieves comparable cost-effectiveness to \runtime.  However,
determining the best way to leverage partial memory is also a search problem,
which we solve elegantly in \runtime.

\subsection{Large-Scale Training}
\label{sec:large-scale-eval}

\begin{figure}[!t]
\centering
    \includegraphics[width=0.95\linewidth]{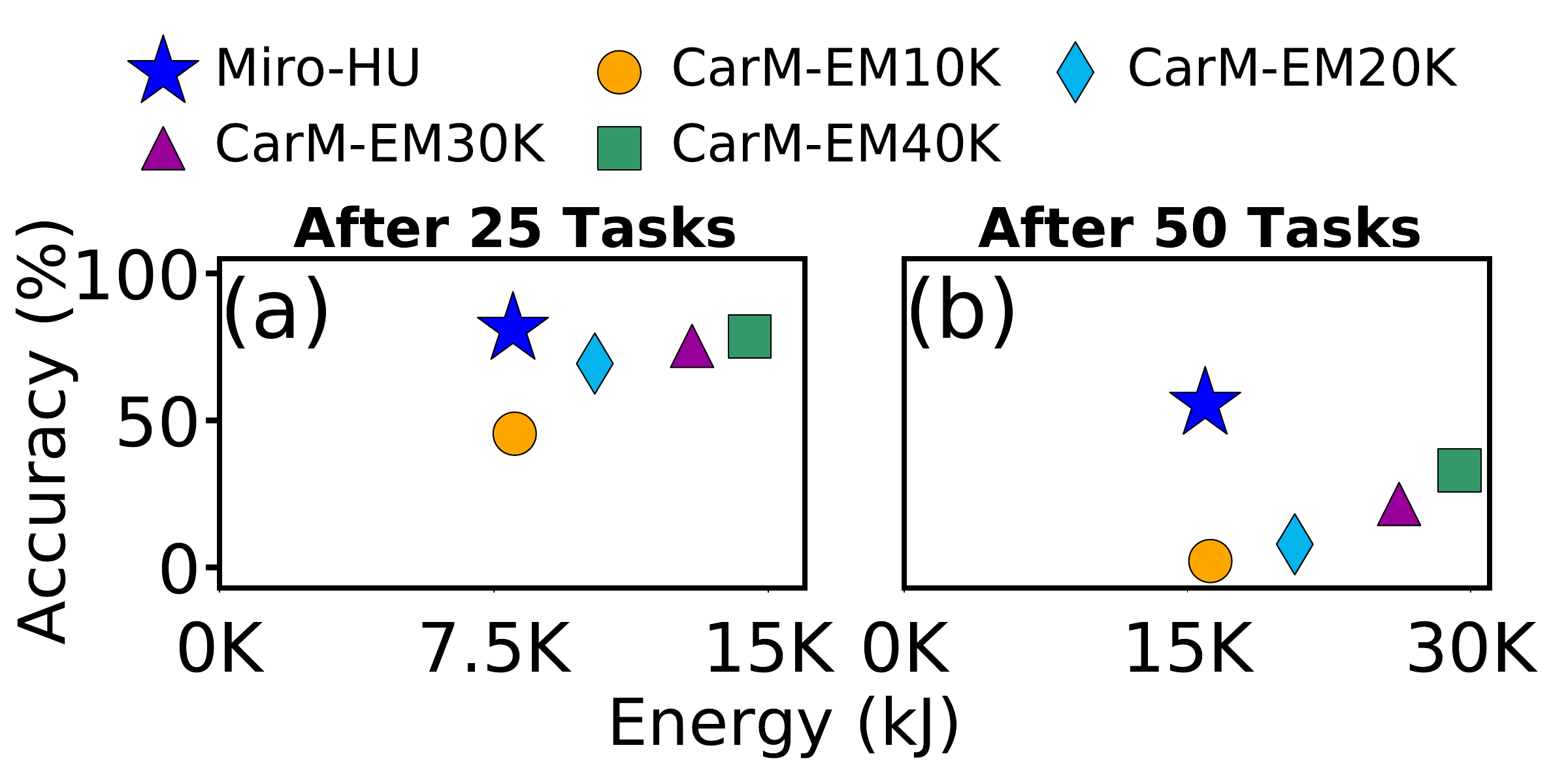}
    \caption{Energy-accuracy trade-offs after completing 25 tasks (a) and 50 tasks (b) for ImageNet1k-50Tasks.}
\label{fig:imagenet1k}
\vspace{-0.2in}
\end{figure}

In prior CL works,
ImageNet1k has commonly been evaluated as a sequence of
10 tasks,
each consisting of 100 classes with 1,300 samples per class.
However, longer task sequences better represent realistic use cases
where edge devices require frequent, real-time model updates. Thus, our
evaluation with ImageNet1k is based on 50 tasks, with each task including 20
classes. Due to the prohibitive cost of fine-grained exploration in \bsfull and
\bhfull, we focus on comparing \runtime with four static strategies based on
\carm with varying EM sizes: 10K (\carm-\textsf{EM10K}) to 40K
(\carm-\textsf{EM40K}) samples, which cover the energy-accuracy trade-off
spectrum. For \runtime, we use
the highest utility (\textsf{HU}) as the conf selection metric.

To highlight how \runtime behaves in the lengthy task sequence, we present
energy and accuracy after completing 25 and 50 tasks
in~\autoref{fig:imagenet1k}. \runtime consistently outperforms all four static
confs in terms of cost-effectiveness in both task completion points.
Importantly, the performance gap between \runtime and \carm widens as more
tasks are trained. For instance, \runtime and \carm-\textsf{EM20K} maintain
similar utility until 10 around tasks, but after 25 tasks, \runtime exhibits
45.7\% higher utility, which further increases to 218.63\% after 50 tasks.
This is expected because static memory confs cannot adapt to the varying task
properties, which become more evident in longer task sequences.

\subsection{Design Validation}

For design validation, we utilize \runtime-\textsf{HU} for ResNet-32 training
on CIFAR100 with a memory budget of 25K samples.

\begin{figure}
    \centering
    \includegraphics[width=\linewidth]{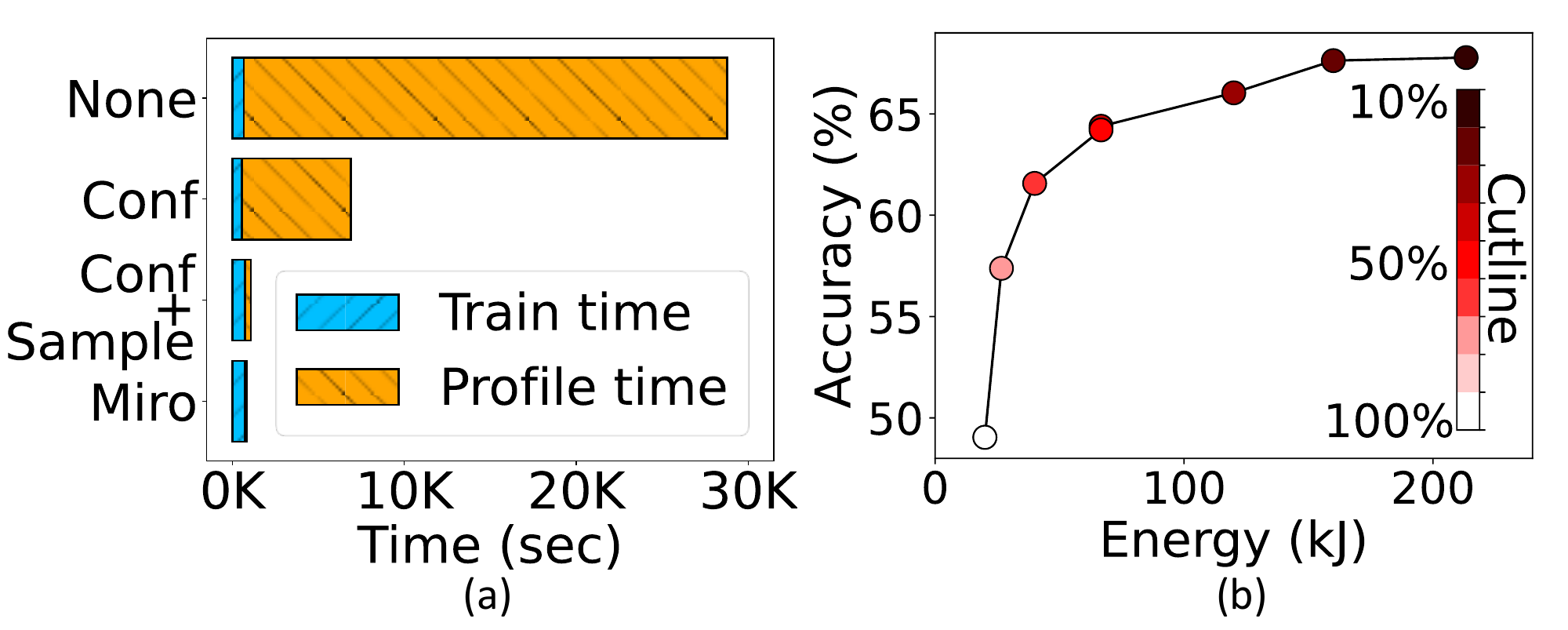}
    \caption{(a) Time breakdown for various profilers with and w/o cost-reduction optimizations. (b) The memory-accuracy trade-off results from different cutline values.}
    \label{fig:cutline}
    \vspace{-0.2in}
\end{figure}

\parlabel{Profiler Optimizations.}
We compare four profiler variations to assess cost-reduction techniques.
(1) \textsf{None} performs exhaustive search and full profiling without any
cost reduction. (2) \textsf{Conf} reduces the number of confs to explore
to 20\% (14 confs). (3) \textsf{Conf+Sample} adds sub-sampling of training
data (5\%) on top of \textsf{Conf} while still profiling full epochs. The
results in~\autoref{fig:cutline}(a) demonstrate
profiling time and model training time, which combined constitute the
end-to-end time. We make two observations. First, profiling costs decrease as
we add cost-reduction techniques one by one, making the profiling cost of
\runtime 346$\times$ faster than \textsf{None}. Second, the confs chosen from
these profilers do not significantly increase
model training time, making \runtime the most effective in balancing
profiling time \vs model training time. Hence, our main features greatly
contribute to \runtime's system performance. Furthermore, we observe that all
profilers achieve competitive accuracies (62.07--63.48\%), suggesting that
other profilers do not offer better cost-effectiveness.

\parlabel{Checkpoint.}
Using checkpoint avoids collecting unreliable profiling results during the
earlier epochs, where spikes and noises in the training loss appear.
To confirm the effectiveness of this approach, we compare our method to the
method without checkpoint: \textsf{Epoch-15}, which runs 15 epochs from epoch
zero. Note that our method takes a checkpoint at epoch 10 and runs each conf
for 5 epochs to finish it at epoch 15.
We compare profiling time and accuracy. Our method achieves comparable
accuracy (0.26\% less) to \textsf{Epoch-15} while reducing profiling costs as
much as 70\%.

\parlabel{Cutline.}
To properly filter out unpromising confs, \runtime relies on an appropriate
cutline value. Thus, we evaluate a wide range of cutline values from 10\% to
100\% (\ie, no cutline)
and present the energy-accuracy trade-off results in~\autoref{fig:cutline}(b).
A cutline of 80--100\% leaves too many unpromising confs, increasing the risk
of accidentally selecting a conf with low accuracy but high utility due to
very small memory usage
(as shown in our example in~\autoref{fig:illustrative_example}). We found
that setting the cutline between 30--50\% works well for our benchmarks,
while \runtime could further improve its adaptability by adjusting the
cutline based on the current accuracy distribution.
\section{Discussion}
\label{sec:discussion}

\parlabel{Storage Lifespan and Capacity.}
MLC-based SD cards today endure around 10,000 write-and-erase cycles.
\runtime writes samples to storage only once per task, with later
data-swapping operations involving only reads. Thus, with proper wear
leveling, \runtime will not significantly harm the storage lifespan.

However, 
when the number of tasks scales up,
realistic HEM-based on-device
CL may exceed the storage capacity. In such cases, 
one option is to embrace larger external storage in edge or cloud
servers, which may require a higher level of data
privacy. The other option is to evict some old samples in storage
to make room for 
new data, 
which may require a 
data sampling strategy. 
In either case, a good data selection strategy for a
deeper memory hierarchy will be needed.

\parlabel{Temporal Data Locality.}
Consider video analytics as an example. On-road or urban videos have high temporal data locality, requiring NN models to focus on recent object appearances: unrelated objects may not be the
target of inference. In other words, the NN models may need to be tailored
to a specific object distribution at any time.
To enhance the
effectiveness of HEM for this particular scenario, we can allow it to use
in-storage data collected from recent tasks more aggressively by increasing
their swapping ratio and EM space. To do this, \runtime needs to be
improved with 
task-level resource allocation policies for
both memory and I/O.

\parlabel{Profiling.}
Our evaluation datasets are structured to reveal data drift between two
adjacent tasks, making performing a search each time a task is added a
natural choice. Nonetheless, there may be more efficient approaches for
certain workloads like video analytics, where consecutive tasks defined by
fixed time windows often exhibit high temporal locality and similar class
distributions~\cite{ekya, recl}. In such cases, it may be possible to treat
two adjacent tasks as a single, larger task that does not require parameter
reconfiguration.

\parlabel{Energy-efficient Training Iterations.}
Most of the energy consumption in on-device CL comes from GPUs that execute
training iterations. This highlights the potential effectiveness of
strategies like layer skipping~\cite{Huang2016DeepNW}, layer
freezing~\cite{yang2023efficientsc}, and sparsity-aware training~\cite{spacenet, sparcl} in improving the energy bottleneck of GPUs by
reducing the computational costs of training iterations. These optimization
techniques are complementary to \runtime's approach, which enhances energy
efficiency by reducing the total number of SB and EM samples required for
training. By combining both approaches, we can make on-device CL even more
energy-efficient.

\section{Related Work}

One of the most realistic CL scenarios is class-incremental learning (CIL),
which involves inference without task IDs~\cite{gepperthH16}. We focus on
EM-based methods for our study since they tend to show better accuracies in
CIL~\cite{gdumb}.

\parlabel{Episodic Memory Management.}
Several works aim to improve the quality of EM samples through various
methods, such as storing samples that represent the mean and boundary of each
class distribution~\cite{mnemonics}, generating a
coreset~\cite{Borsos2020CoresetsVB}, perturbing styles of remembered samples
using GANs~\cite{Cong2020GANMW}, and promoting the diversity of samples in
EM~\cite{rainbow}. These methods may involve excessive computation or
difficulty~\cite{Cong2020GANMW,Borsos2020CoresetsVB}. However,
most of these methods show only marginal accuracy improvements over uniform
random sampling~\cite{rwalk,castro2018eccv,icarl}. Some works propose to
directly generate samples of past
tasks~\cite{Shin2017ContinualLW,Seff2017ContinualLI,Wu2018MemoryRG,hu2018overcoming}.
Unlike these works addressing sampling efficiency, the focus of this study is
on devising system-oriented techniques that smartly manage samples across the
memory-storage hierarchy.

Recently, BudgetCL~\cite{budgetcl} introduced a single-level EM architecture
based on storage.
It utilizes uniform sampling from EM to limit training within a predefined
computational budget. While BudgetCL can serve as a replacement for HEM, it
may face training slowdowns due to frequent I/O operations, as demonstrated
by the CarM work~\cite{carm}. Nevertheless, \runtime can be adapted to work
with BudgetCL's memory architecture for cost-effectiveness since the
variation in trained EM samples in BudgetCL corresponds to data swapping
within the storage. This results in minimal impact on the system's energy
consumption, as inferred by ~\autoref{fig:power_consumptions}.

\parlabel{Continual Learning on the Edge.}
There are a few noteworthy works done for improving CL on the edge.
Hayes~\etal~\cite{hayes2022online} explore seven online CL methods that train
one sample at a time using CNN models designed for embedded devices,
providing algorithmic guidance in this area.
Kwon~\etal~\cite{kwon2021exploring} compare the trade-offs between
performance, storage, and compute and memory costs for regularization \vs EM
methods. They observe that EM methods like iCaRL~\cite{icarl} offer the best
performance trade-offs.
Some of their claims emphasize the need for system approaches to improve EM
methods, which is aligned with our research goal.

\parlabel{Offline Learning on the Edge.}
Several recent system studies also focus on the memory and energy efficiency
of on-device learning under conventional non-CL setups. Sage~\cite{sage}
performs operator- and graph-level optimizations at compile time and memory
management, such as gradient checkpointing~\cite{checkpointing}, at runtime
for memory-efficient training. Melon~\cite{melon} optimizes model training
performance under a given memory capacity by leveraging
micro-batch~\cite{microbatch} and recomputation~\cite{recomputation}
techniques. E$^2$-Train~\cite{e2train} cares about both training time and
energy cost. To achieve the goal, it applies stochastic mini-batch dropping,
selective layer update, and sign prediction for low-cost backpropagation.
Offline on-device training is considered a prerequisite for
continual learning. Unlike these studies, we tackle practical on-device
CL directly.

\section{Conclusion}

In this work, we investigate the design space of HEM to share insights on
enabling cost-effective on-device CL. To achieve this, we introduce \runtime,
a system runtime that optimizes HEM for cost-effectiveness under various HW
resource constraints. \runtime dynamically reconfigures HEM through runtime
profiling to adapt to the current learning conditions rather than relying on
simple heuristics or past decisions. We evaluate the practicality of \runtime
using various CL tasks.

\parlabel{Acknowledgments:}
We thank our shepherd, anonymous reviewers, Kwangin Kim, Dongyoon Ryu, Taeyoon Kim, and Heelim Hong for their insightful comments and feedback. This work was supported by the 2023 Research Fund (1.230019) of UNIST, ETRI grant [23ZS1300], and the IITP grants (No.2021-0-02068, AI Innovation Hub 5\%, 2022-0-00113 15\%) funded by the Korea government (MSIT).

\clearpage
\newpage
\balance
\bibliographystyle{abbrv}
\bibliography{bibliography}

\begin{thebibliography}{10}

\bibitem{carm_code}
{Carousel Memory (CarM) - Official Pytorch Implementation}.
\newblock \url{https://github.com/supersoob/CarM}.

\bibitem{power_code}
{Convenient Power Measurements on the Jetson TX2/Tegra X2 Board}.
\newblock
  \url{https://embeddeddl.wordpress.com/2018/04/25/convenient-power-measurements-on-the-jetson-tx2-tegra-x2-board/}.

\bibitem{jetson_tx2}
{Jetson TX2 Module}.
\newblock \url{https://developer.nvidia.com/embedded/jetson-tx2}.

\bibitem{jetson_xavier_nx}
{Jetson Xavier NX Series}.
\newblock
  \url{https://www.nvidia.com/en-us/autonomous-machines/embedded-systems/jetson-xavier-nx}.

\bibitem{power_manual}
{OEM PRODUCT DESIGN GUIDE: NVIDIA Jetson TX2}.
\newblock
  \url{https://usermanual.wiki/Pdf/jetsontx2oemproductdesignguide.2134990230.pdf}.

\bibitem{sharedmem}
{Python Shared Memory}.
\newblock \url{https://docs.python.org/3/library/multiprocessing.html}.

\bibitem{TinyImageNet}
{Tiny ImageNet Visual Recognition Challenge}.
\newblock \url{https://tiny-imagenet.herokuapp.com}.

\bibitem{rainbow}
J.~Bang, H.~Kim, Y.~Yoo, J.-W. Ha, and J.~Choi.
\newblock {Rainbow Memory: Continual Learning with a Memory of Diverse
  Samples}.
\newblock In {\em CVPR}, 2021.

\bibitem{dsads}
B.~Barshan and K.~Altun.
\newblock {Daily and Sports Activities}.
\newblock UCI Machine Learning Repository, 2013.
\newblock {DOI}: https://doi.org/10.24432/C5C59F.

\bibitem{ekya}
R.~Bhardwaj, Z.~Xia, G.~Ananthanarayanan, J.~Jiang, N.~Karianakis, Y.~Shu,
  K.~Hsieh, V.~Bahl, and I.~Stoica.
\newblock {Ekya: Continuous Learning of Video Analytics Models on Edge Compute
  Servers}.
\newblock In {\em NSDI}, 2022.

\bibitem{Borsos2020CoresetsVB}
Z.~Borsos, M.~Mutn{\`y}, and A.~Krause.
\newblock {Coresets via Bilevel Optimization for Continual Learning and
  Streaming}.
\newblock In {\em NeurIPS}, 2020.

\bibitem{darker}
P.~Buzzega, M.~Boschini, A.~Porrello, D.~Abati, and S.~CALDERARA.
\newblock {Dark Experience for General Continual Learning: a Strong, Simple
  Baseline}.
\newblock In {\em NeurIPS}, 2020.

\bibitem{castro2018eccv}
F.~M. Castro, M.~J. Marin-Jimenez, N.~Guil, C.~Schmid, and K.~Alahari.
\newblock {End-to-End Incremental Learning}.
\newblock In {\em ECCV}, 2018.

\bibitem{rwalk}
A.~Chaudhry, P.~K. Dokania, T.~Ajanthan, and P.~H.~S. Torr.
\newblock {Riemannian Walk for Incremental Learning: Understanding Forgetting
  and Intransigence}.
\newblock In {\em ECCV}, 2018.

\bibitem{AGEM}
A.~Chaudhry, M.~Ranzato, M.~Rohrbach, and M.~Elhoseiny.
\newblock {Efficient Lifelong Learning with A-GEM}.
\newblock In {\em ICLR}, 2019.

\bibitem{tiny}
A.~Chaudhry, M.~Rohrbach, M.~Elhoseiny, T.~Ajanthan, P.~K. Dokania, P.~H. Torr,
  and M.~Ranzato.
\newblock {On Tiny Episodic Memories in Continual Learning}.
\newblock {\em arXiv:1902.10486}, 2019.

\bibitem{chauhan2020contauth}
J.~Chauhan, Y.~D. Kwon, P.~Hui, and C.~Mascolo.
\newblock {ContAuth: Continual Learning Framework for Behavioral-based User
  Authentication}.
\newblock {\em IMWUT}, 4(4):1--23, 2020.

\bibitem{recomputation}
T.~Chen, B.~Xu, C.~Zhang, and C.~Guestrin.
\newblock {Training Deep Nets with Sublinear Memory Cost}.
\newblock {\em arXiv preprint arXiv:1604.06174}, 2016.

\bibitem{Cong2020GANMW}
Y.~Cong, M.~Zhao, J.~Li, S.~Wang, and L.~Carin.
\newblock {GAN Memory with No Forgetting}.
\newblock In {\em NeurIPS}, 2020.

\bibitem{diwan2022continual}
A.~Diwan, C.-F. Yeh, W.-N. Hsu, P.~Tomasello, E.~Choi, D.~Harwath, and
  A.~Mohamed.
\newblock {Continual Learning for On-Device Speech Recognition using
  Disentangled Conformers}.
\newblock {\em arXiv preprint arXiv:2212.01393}, 2022.

\bibitem{tcp_reno}
K.~Fall and S.~Floyd.
\newblock {Simulation-Based Comparisons of Tahoe, Reno and SACK TCP}.
\newblock {\em SIGCOMM Comput. Commun. Rev.}, 26(3):5–21, jul 1996.

\bibitem{gepperthH16}
A.~Gepperth and B.~Hammer.
\newblock {Incremental Learning Algorithms and Applications}.
\newblock In {\em {ESANN}}, 2016.

\bibitem{sage}
I.~Gim and J.~Ko.
\newblock {Memory-Efficient DNN Training on Mobile Devices}.
\newblock In {\em MobiSys}, 2022.

\bibitem{checkpointing}
A.~Griewank and A.~Walther.
\newblock {Algorithm 799: Revolve: An Implementation of Checkpointing for the
  Reverse or Adjoint Mode of Computational Differentiation}.
\newblock {\em ACM Trans. Math. Softw.}, 26(1):19–45, mar 2000.

\bibitem{tcp_cubic}
S.~Ha, I.~Rhee, and L.~Xu.
\newblock {CUBIC: A New TCP-Friendly High-Speed TCP Variant}.
\newblock {\em SIGOPS Oper. Syst. Rev.}, 42(5):64–74, jul 2008.

\bibitem{hayes2022online}
T.~L. Hayes and C.~Kanan.
\newblock {Online Continual Learning for Embedded Devices}.
\newblock In {\em Conference on Lifelong Learning Agents}, 2022.

\bibitem{resnet}
K.~He, X.~Zhang, S.~Ren, and J.~Sun.
\newblock {Deep Residual Learning for Image Recognition}.
\newblock In {\em 2016 IEEE Conference on Computer Vision and Pattern
  Recognition (CVPR)}, pages 770--778, 2016.

\bibitem{hu2018overcoming}
W.~Hu, Z.~Lin, B.~Liu, C.~Tao, Z.~Tao, J.~Ma, D.~Zhao, and R.~Yan.
\newblock {Overcoming Catastrophic Forgetting via Model Adaptation}.
\newblock In {\em ICLR}, 2019.

\bibitem{Huang2016DeepNW}
G.~Huang, Y.~Sun, Z.~Liu, D.~Sedra, and K.~Q. Weinberger.
\newblock {Deep Networks with Stochastic Depth}.
\newblock In {\em European Conference on Computer Vision}, 2016.

\bibitem{microbatch}
Y.~Huang, Y.~Cheng, A.~Bapna, O.~Firat, M.~X. Chen, D.~Chen, H.~Lee, J.~Ngiam,
  Q.~V. Le, Y.~Wu, and Z.~Chen.
\newblock {GPipe: Efficient Training of Giant Neural Networks Using Pipeline
  Parallelism}.
\newblock In {\em NeurIPS}, 2019.

\bibitem{recl}
M.~Khani, G.~Ananthanarayanan, K.~Hsieh, J.~Jiang, R.~Netravali, Y.~Shu,
  M.~Alizadeh, and V.~Bahl.
\newblock {RECL: Responsive Resource-Efficient Continuous Learning for Video
  Analytics}.
\newblock In {\em NSDI}, 2023.

\bibitem{cifar}
A.~Krizhevsky, G.~Hinton, et~al.
\newblock {Learning Multiple Layers of Features from Tiny Images}.
\newblock 2009.

\bibitem{kwon2021exploring}
Y.~D. Kwon, J.~Chauhan, A.~Kumar, P.~H. HKUST, and C.~Mascolo.
\newblock {Exploring System Performance of Continual Learning for Mobile and
  Embedded Sensing Applications}.
\newblock In {\em IEEE/ACM SEC}, 2021.

\bibitem{carm}
S.~Lee, M.~Weerakoon, J.~Choi, M.~Zhang, D.~Wang, and M.~Jeon.
\newblock {CarM: Hierarchical Episodic Memory for Continual Learning}.
\newblock In {\em DAC}, 2022.

\bibitem{leite2022resource}
C.~F.~S. Leite and Y.~Xiao.
\newblock {Resource-Efficient Continual Learning for Sensor-Based Human
  Activity Recognition}.
\newblock {\em ACM Transactions on Embedded Computing Systems}, 21(6):1--25,
  2022.

\bibitem{mnemonics}
Y.~Liu, Y.~Su, A.-A. Liu, B.~Schiele, and Q.~Sun.
\newblock {Mnemonics Training: Multi-Class Incremental Learning Without
  Forgetting}.
\newblock In {\em IEEE/CVF Conference on Computer Vision and Pattern
  Recognition (CVPR)}, June 2020.

\bibitem{LopezPaz2017GradientEM}
D.~Lopez-Paz and M.~Ranzato.
\newblock {Gradient Episodic Memory for Continual Learning}.
\newblock In {\em NeurIPS}, 2017.

\bibitem{mccloskeyC89}
M.~McCloskey and Neal.
\newblock {Catastrophic Interference in Connectionist Networks: The Sequential
  Learning Problem}.
\newblock In {\em Psychology of Learning and Motivation}, 1989.

\bibitem{pellegrini2021continual}
L.~Pellegrini, V.~Lomonaco, G.~Graffieti, and D.~Maltoni.
\newblock {Continual Learning at the Edge: Real-Time Training on Smartphone
  Devices}.
\newblock {\em arXiv preprint arXiv:2105.13127}, 2021.

\bibitem{budgetcl}
A.~Prabhu, H.~A. Al~Kader~Hammoud, P.~K. Dokania, P.~H. Torr, S.-N. Lim,
  B.~Ghanem, and A.~Bibi.
\newblock {Computationally Budgeted Continual Learning: What Does Matter?}
\newblock In {\em Proceedings of the IEEE/CVF Conference on Computer Vision and
  Pattern Recognition (CVPR)}, pages 3698--3707, June 2023.

\bibitem{gdumb}
A.~Prabhu, P.~H. Torr, and P.~K. Dokania.
\newblock {GDumb: A Simple Approach that Questions Our Progress in Continual
  Learning}.
\newblock In {\em ECCV}, 2020.

\bibitem{icarl}
S.-A. Rebuffi, A.~Kolesnikov, G.~Sperl, and C.~H. Lampert.
\newblock {iCaRL: Incremental Classifier and Representation Learning}.
\newblock In {\em CVPR}, 2017.

\bibitem{ercl}
D.~Rolnick, A.~Ahuja, J.~Schwarz, T.~P. Lillicrap, and G.~Wayne.
\newblock {Experience Replay for Continual Learning}.
\newblock 2019.

\bibitem{imagenet_1k}
O.~Russakovsky, J.~Deng, H.~Su, J.~Krause, S.~Satheesh, S.~Ma, Z.~Huang,
  A.~Karpathy, A.~Khosla, M.~Bernstein, et~al.
\newblock {ImageNet Large Scale Visual Recognition Challenge}.
\newblock {\em Int J Comput Vis}, 115:211--252, 2015.

\bibitem{Salamon2016DeepCN}
J.~Salamon and J.~P. Bello.
\newblock {Deep Convolutional Neural Networks and Data Augmentation for
  Environmental Sound Classification}.
\newblock {\em IEEE Signal Processing Letters}, 24:279--283, 2016.

\bibitem{urbansound}
J.~Salamon, C.~Jacoby, and J.~P. Bello.
\newblock {A Dataset and Taxonomy for Urban Sound Research}.
\newblock In {\em 22nd {ACM} International Conference on Multimedia
  (ACM-MM'14)}, pages 1041--1044, Orlando, FL, USA, Nov. 2014.

\bibitem{schiemer4357622online}
M.~Schiemer, L.~Fang, S.~Dobson, and J.~Ye.
\newblock {Online Continual Learning for Human Activity Recognition}.
\newblock {\em Available at SSRN 4357622}.

\bibitem{Seff2017ContinualLI}
A.~Seff, A.~Beatson, D.~Suo, and H.~Liu.
\newblock {Continual Learning in Generative Adversarial Nets}.
\newblock {\em arXiv preprint arxiv:1705.08395}, 2017.

\bibitem{shaheen2022continual}
K.~Shaheen, M.~A. Hanif, O.~Hasan, and M.~Shafique.
\newblock {Continual Learning for Real-World Autonomous Systems: Algorithms,
  Challenges and Frameworks}.
\newblock {\em Journal of Intelligent \& Robotic Systems}, 105(1):9, 2022.

\bibitem{aser}
D.~Shim, Z.~Mai, J.~Jeong, S.~Sanner, H.~Kim, and J.~Jang.
\newblock {Online Class-Incremental Continual Learning with Adversarial Shapley
  Value}.
\newblock In {\em AAAI}, 2021.

\bibitem{Shin2017ContinualLW}
H.~Shin, J.~K. Lee, J.~Kim, and J.~Kim.
\newblock {Continual Learning with Deep Generative Replay}.
\newblock In {\em NeurIPS}, 2017.

\bibitem{spacenet}
G.~Sokar, D.~C. Mocanu, and M.~Pechenizkiy.
\newblock {SpaceNet: Make Free Space For Continual Learning}.
\newblock {\em Neurocomputing}, 439:1--11, 2020.

\bibitem{vokinger2021continual}
K.~N. Vokinger, S.~Feuerriegel, and A.~S. Kesselheim.
\newblock {Continual Learning in Medical Devices: FDA’s Action Plan and
  Beyond}.
\newblock {\em The Lancet Digital Health}, 3(6):e337--e338, 2021.

\bibitem{melon}
Q.~Wang, M.~Xu, C.~Jin, X.~Dong, J.~Yuan, X.~Jin, G.~Huang, Y.~Liu, and X.~Liu.
\newblock {Melon: Breaking the Memory Wall for Resource-Efficient on-Device
  Machine Learning}.
\newblock In {\em MobiSys}, 2022.

\bibitem{e2train}
Y.~Wang, Z.~Jiang, X.~Chen, P.~Xu, Y.~Zhao, Y.~Lin, and Z.~Wang.
\newblock {E2-Train: Training State-of-the-art CNNs with Over 80\% Energy
  Savings}.
\newblock In {\em NeurIPS}, 2019.

\bibitem{sparcl}
Z.~Wang, Z.~Zhan, Y.~Gong, G.~Yuan, W.~Niu, T.~Jian, B.~Ren, S.~Ioannidis,
  Y.~Wang, and J.~Dy.
\newblock {SparCL: Sparse Continual Learning on the Edge}.
\newblock In {\em NeurIPS}, 2022.

\bibitem{Wu2018MemoryRG}
C.~Wu, L.~Herranz, X.~Liu, y.~wang, J.~van~de Weijer, and B.~Raducanu.
\newblock {Memory Replay GANs: Learning to Generate New Categories without
  Forgetting}.
\newblock In {\em Advances in Neural Information Processing Systems},
  volume~31. Curran Associates, Inc., 2018.

\bibitem{bic}
Y.~Wu, Y.~Chen, L.~Wang, Y.~Ye, Z.~Liu, Y.~Guo, and Y.~Fu.
\newblock {Large Scale Incremental Learning}.
\newblock In {\em CVPR}, 2019.

\bibitem{yang2023efficientsc}
L.~Yang, S.~Lin, F.~Zhang, J.~Zhang, and D.~Fan.
\newblock {Efficient Self-supervised Continual Learning with Progressive
  Task-correlated Layer Freezing}.
\newblock {\em ArXiv}, abs/2303.07477, 2023.

\bibitem{zeus}
J.~You, J.-W. Chung, and M.~Chowdhury.
\newblock {Zeus: Understanding and Optimizing GPU Energy Consumption of DNN
  Training}.
\newblock In {\em NSDI}, 2023.

\bibitem{e-domainil}
Y.~Zhao, D.~Saxena, and J.~Cao.
\newblock {Memory-Efficient Domain Incremental Learning for Internet of
  Things}.
\newblock In {\em SenSys}, 2023.

\end{thebibliography}


\end{document}